\documentclass[journal,onecolumn,draftclsnofoot,12pt]{IEEEtran}
	\usepackage{graphicx,cite,amsmath,amssymb,autobreak,hhline,booktabs,stfloats,bm,amsthm,hyperref,times,subfigure,multirow,algpseudocode,}
	\usepackage{algorithm,algpseudocode,multicol,threeparttable,dcolumn,latexsym,rotating,textcomp,colortbl,enumerate}
	\usepackage{verbatim}
	\usepackage{caption}
	\usepackage[top=2cm, bottom=2cm, left=2cm, right=2cm]{geometry}
	\usepackage{graphicx}
	\usepackage{grffile}
	\usepackage{stfloats}
\begin{document}

	\newtheorem{lemma}{\bf Lemma }
	\newtheorem{theorem}{\bf Theorem}
	\newtheorem{corollary}{Corollary}
	\allowdisplaybreaks[4]
	\newtheorem{remark}{Remark}
	\newtheorem{assumption}{Assumption}
	\title{Over-the-Air Federated Averaging with Limited Power and Privacy Budgets}
		
	
	\author{Na Yan, Kezhi Wang, Cunhua Pan, Kok Keong Chai, Feng Shu, \\and Jiangzhou Wang, \emph{Fellow, IEEE} \thanks{Part of this work will be presented in IEEE International Conference on Communications (ICC), 28 May – 01 June 2023, Rome, Italy. This work of Na Yan was supported by China Scholarship Council. \itshape (Corresponding author: Kezhi Wang and Cunhua Pan.). \upshape
			Na Yan and Kok Keong Chai are with School of Electronic Engineering and Computer Science, Queen Mary University of London, London E1 4NS, U.K. (e-mail: n.yan, michael.chai@qmul.ac.uk).	
			Kezhi Wang is with Department of Computer Science, Brunel University London, Uxbridge, Middlesex, UB8 3PH, U.K. (email: kezhi.wang@brunel.ac.uk).
			Cunhua Pan is with the National Mobile Communications Research Laboratory, Southeast University, Nanjing 210096, China (email: cpan@seu.edu.cn).
			Feng Shu is with the School of Information and Communication Engineering, Hainan University, Haikou 570228, China, and also with the School of Electronic and Optical Engineering, Nanjing University of Science and Technology, Nanjing 210094, China (e-mail: shufeng0101@163.com).
			Jiangzhou Wang is with the School of Engineering, University of Kent,
			Canterbury CT2 7NT, U.K. (Email: j.z.wang@kent.ac.uk).}
	}
	\maketitle
\vspace{-25pt}
	\begin{abstract}
To jointly overcome the communication bottleneck and privacy leakage of wireless federated learning (FL), this paper studies a differentially private over-the-air federated averaging (DP-OTA-FedAvg) system with a limited sum power budget. 
With DP-OTA-FedAvg, the gradients are aligned by an alignment coefficient and aggregated over the air, and channel noise is employed to protect privacy.
We aim to improve the learning performance by jointly designing the device scheduling, alignment coefficient, and the number of aggregation rounds of federated averaging (FedAvg) subject to sum power and privacy constraints. 
We first present the privacy analysis based on differential privacy (DP) to quantify the impact of the alignment coefficient on privacy preservation in each communication round.
Furthermore, to study how the device scheduling, alignment coefficient, and the number of the global aggregation affect the learning process, we conduct the convergence analysis of DP-OTA-FedAvg in the cases of convex and non-convex loss functions. 
Based on these analytical results, we formulate an optimization problem to minimize the optimality gap of the DP-OTA-FedAvg subject to limited sum power and privacy budgets. 
The problem is solved by decoupling it into two sub-problems. 
Given the number of communication rounds, we conclude the relationship between the number of scheduled devices and the alignment coefficient, which offers a set of potential optimal solution pairs of device scheduling and the alignment coefficient. Thanks to the reduced search space, the optimal solution can be efficiently obtained. The effectiveness of the proposed policy is validated through simulations.
	\centerline{\textbf{Intex Terms} } 
	 Federated averaging, differential privacy, over-the-air computation, and device scheduling.
	\end{abstract}
	
	\section{Introduction}
	
With the rapid increase in data volume and computing capability of edge devices, artificial intelligence (AI) and Internet of Things (IoT) are well-developed as a result of the unprecedented success of machine learning (ML) techniques, especially deep learning\cite{yang2020federated}. 
These systems normally employ highly parameterized models, such as deep neural networks (DNNs), which are trained by the massive data samples generated or collected by edge devices, e.g. smartphones and sensors. 
The conventional strategy for training these models is to aggregate all these raw data to a central server with high computing capability, where the training is performed \cite{chen2019deep}.
However, such a centralized training paradigm is becoming more and more costly due to the transmission of raw samples with the dramatic growth in data amount. Furthermore, the raw data usually contains some personal information, and thus the users may refuse to share them with the server. All the above reasons inspire the development of federated learning (FL), which is a kind of privacy-preserving distributed ML paradigm \cite{mcmahan2017communication,liu2020federated,yang2022federated}.
	
FL enables the devices to train models collaboratively with the help of a central controller, such as a base station (BS) \cite{mcmahan2017communication,liu2020federated,yang2022federated}. Instead of uploading the raw data to the BS, the model parameters and the gradients are exchanged between the devices and the BS. By training models locally, FL not only makes full use of the computing capability of the edge devices, but also effectively reduces the power consumption, latency, and privacy exposure caused by the transmission of the massive datasets. However, despite these promising benefits, FL still involves the following challenges. First, FL suffers from communication bottlenecks due to the high dimension of each local update, especially when a large number of participants try to upload gradients via a resource-limited wireless multiple access channel (MAC). This also leads to considerable upload latency as the bandwidth allocated to each participant decreases with the increased number of devices	 \cite{kairouz2021advances,mcmahan2017communication}.
Second, although FL offers basic privacy protection, which benefits from the fact that all raw data is processed locally, it is far from sufficiency if some attacks are applied to the exchanged messages, i.e., the gradients \cite{melis2019exploiting,nasr2018comprehensive}. This is because the gradients are obtained based on local data and therefore may contain some information of raw data \cite{song2017machine}.
	
One promising countermeasure to jointly overcome the two challenges is over-the-air FL (OTA-FL) \cite{nazer2007computation,zhu2019broadband,amiri2020machine} with differential privacy (DP)  \cite{dwork2014algorithmic}, referred to as DP-OTA-FL. 
On one hand, DP \cite{dwork2014algorithmic} prevents privacy leakage of FL by introducing random noise into the disclosed statistics, i.e., gradients or model parameters, to mask the contribution of any individual data point. 
On the other hand, OTA-FL schedules the devices to convey their gradients simultaneously via a shared wireless MAC with analog signals, i.e., without converting the gradients to discrete coded symbols which need to be decoded at the BS. 
Then, the gradients are directly aggregated ``over-the-air" thanks to the waveform-superposition property of a MAC. With OTA-FL, the bandwidth used for transmitting the gradients is independent of the number of devices, which makes it more efficient than the traditional communication-and-computation separation method, especially when the number of devices is large \cite{nazer2007computation,goldenbaum2013harnessing}. 
Therefore, OTA-FL is expected to significantly relieve the above-mentioned communication bottleneck and reduce the communication and computation latency. 
However, a major drawback of such uncoded analog transmission is that the aggregation error originated from the channel fading and noise perturbation degrades the learning performance.

Existing works minimized the aggregation error by means of optimizing the hyper-parameters, such as learning rate \cite{guo2020analog,zhang2021federated}, power control \cite{yu2020optimizing,zhang2021gradient,cao2021optimized,yan2022private} and device selection \cite{ma2021user}. In order to eliminate the fading-related error, some works considered the aligned OTA-FL \cite{zhu2019broadband,sery2021over,seif2020wireless,koda2020differentially,liu2020privacy}, where all the gradients are aligned by a constant, referred to as alignment coefficient, by performing the pre-processing mechanism. In this way, the impact of the fading channel becomes a constant and can be easily removed by performing an inverse operation of the pre-processing at the BS. 
However, the alignment coefficient is limited by the participant with the worst channel condition due to the peak transmit power constraint, which can result in a very low signal-to-noise ratio (SNR), especially in the case that all the devices are scheduled in the training \cite{seif2020wireless,koda2020differentially,liu2020privacy}. To improve the alignment coefficient, the authors of \cite{zhu2019broadband,sery2021over} set a threshold to schedule the devices with better channel qualities to participate in the training. However, the optimal threshold was not given. 

Some works studied the over-the-air federated averaging (OTA-FedAvg) \cite{fan2021joint,zhu2019broadband,cao2021optimized,xia2021fast,xu2021learning} to further reduce the communication cost and the negative impact of the communication on the learning process by performing multiple local training rounds before each global aggregation.
The work of \cite{zhu2019broadband} considered a broadband OTA-FedAvg system and a set of interesting communication-learning tradeoffs were derived.
Subsequently, the joint design of device scheduling and channel-inversion-based power scaling was investigated in \cite{fan2021joint} and a channel state information (CSI) based device selection scheme was proposed in \cite{xia2021fast} to achieve reliable model aggregation. 
Then, by considering multi-antenna OTA-FedAvg systems, a joint device scheduling and receive beamforming design was studied in \cite{yang2020federated}. 
However, all the studies considered a fixed number of aggregation rounds of FedAvg, and most of the works commonly considered the peak transmit power constraint of each device. 
The tradeoff between the reduced transmission disturbance and the increased local training error due to the increased rounds of local training, i.e., the reduced aggregation rounds, is also worth investigating under the sum power constraint, which is important in guiding the design of device scheduling and aggregation of FedAvg.
	
In this paper, a scheme is proposed to jointly design device scheduling, alignment coefficient, and global aggregation for a differentially private OTA-FedAvg (DP-OTA-FedAvg) system with limited sum power and privacy budgets. 
The device scheduling, alignment coefficient, and global aggregation can affect the performance of DP-OTA-FedAvg in two ways. 
On one hand, in each communication round, scheduling more devices to participate in the training is beneficial to alleviate the error of the average gradient. 
However, the alignment coefficient may decrease with the increased number of the scheduled devices as it is more likely to involve the devices with poor channel conditions, which can significantly lower down the alignment coefficient \cite{seif2020wireless}, thus degrading the utility of the aggregated gradient.
Therefore, there is a tradeoff between the number of scheduled devices and the alignment coefficient. It also means that there is an optimal threshold for device scheduling.
Additionally, scheduling more devices in each aggregation round may consume more power. As a result, the number of aggregation rounds will be reduced due to the limited sum power budget. Then, the number of local training will increase with the reduced number of global aggregation rounds, which leads to a larger local training error. Therefore, it is crucial for DP-OTA-FedAvg systems with limited sum power budget to design the device scheduling, alignment coefficient, and aggregation rounds. The main contributions can be summarized as follows:
	\begin{itemize}
		\item We jointly design the device scheduling, alignment coefficient, and the number of aggregation rounds of DP-OTA-FedAvg subject to limited sum power constraint. To the best of our knowledge, this is the first work to investigate the tradeoff between the number of scheduled devices and the alignment coefficient of aligned OTA-FL, and the tradeoff between the aggregation distortion and the local training error with sum power constraint of OTA-FedAvg. 
		
		\item To characterize the impact of the alignment coefficient on the privacy preservation of OTA-FedAvg in each communication round, we first conduct the privacy analysis. Then, we derive the closed-form expressions of the optimality gap and the average-squared gradient to demonstrate the convergence of DP-OTA-FedAvg in the cases of convex and non-convex loss functions, respectively. These closed-form expressions quantify the impact of analog over-the-air aggregation on the convergence of DP-OTA-FedAvg, characterizing how the design of the alignment coefficient, device scheduling, and the number of aggregation rounds can affect the privacy protection and the performance of DP-OTA-FedAvg.
		
		\item Based on these closed-form theoretical results, we formulate an optimization problem to minimize the optimality gap by jointly designing the device scheduling, alignment coefficient, and aggregation rounds considering the limited sum power and privacy budgets. 
		
		\item The problem is decoupled into two sub-problems. By giving the number of communication rounds, the optimal design of device scheduling and alignment coefficient is studied. We obtain limited potential optimal solution pairs by exploring the relationship between the number of scheduled devices and the alignment coefficient. Thanks to the reduced search space, the optimal solution can be efficiently obtained. Given the optimal device scheduling and alignment coefficient, the optimal number of aggregation rounds can be obtained by searching a limited solution space.
	\end{itemize}

\subsection{Organization}
The remainder of this paper is organized as follows. In Section \ref{proj5-section2}, we present the system model, aligned OTA-FedAvg, and the definitions of DP. The theoretically analytical results are presented in Section \ref{proj5-section3}. We formulate an optimization problem in Section \ref{proj5-section4}. The simulation results are shown in Section \ref{proj5-section5} and we conclude the paper in Section \ref{proj5-section6}.
	
	\section{System Model and preliminaries}~\label{proj5-section2}
\begin{figure}
		\centering
		\includegraphics[scale=0.78]{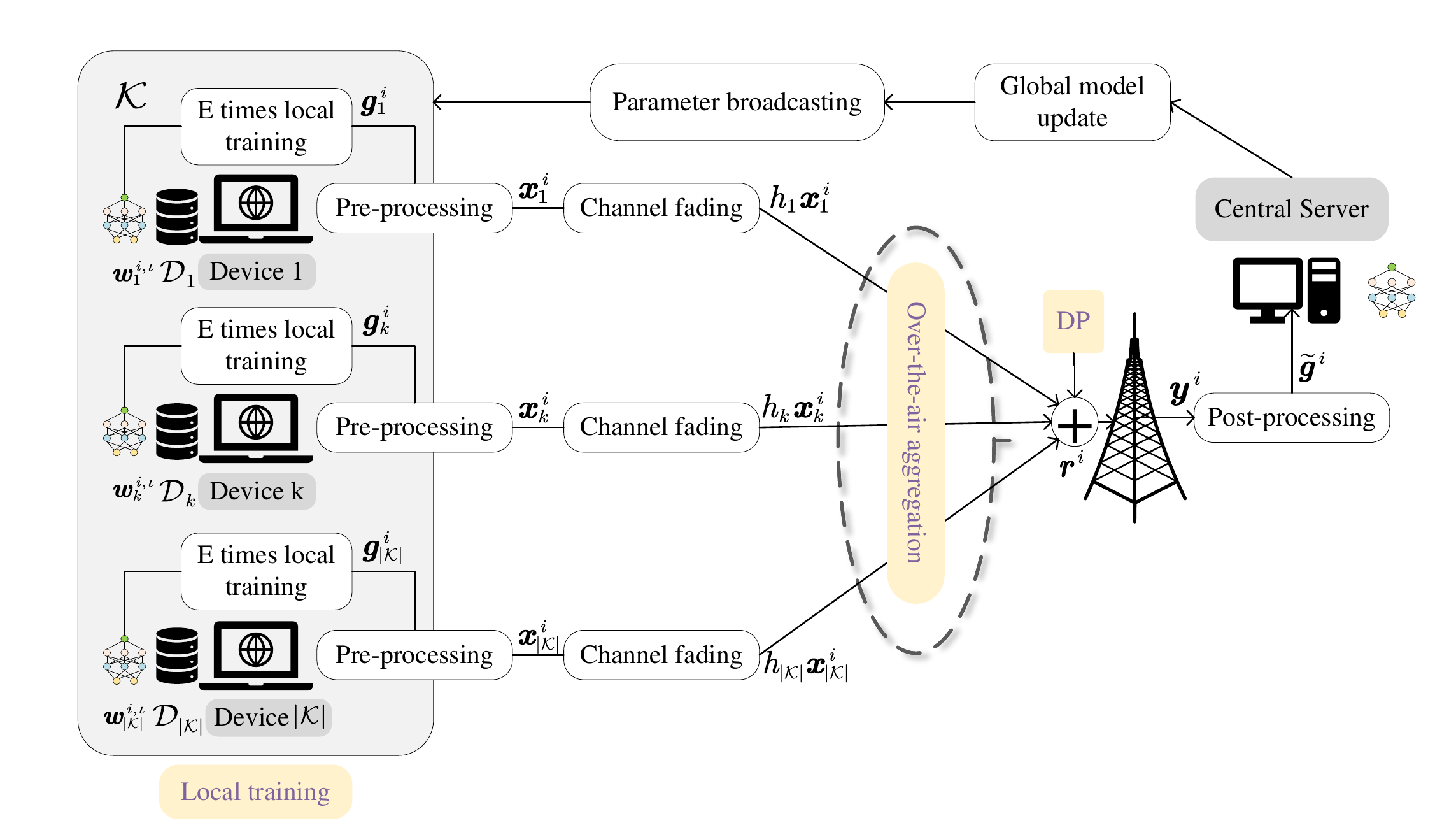}
		\caption{The procedure of DP-OTA-FL.} \label{proj5-fig1}
	\end{figure}

	As shown in Fig. \ref{proj5-fig1},  we consider a DP-OTA-FedAvg system consisting of a BS and $N$ edge devices indexed by $\mathcal{N} =\left \{1, \cdots, N \right \} $. Assume that each device of index $k\in \mathcal{N}$ stores a local dataset $\mathcal{D} _k$ which contains $D_k$ pairs of training samples $\left( \boldsymbol{u},v \right)$ where $\boldsymbol{u}$ is the raw data for training and $v$ is the corresponding label. For simplicity, we assume that $ D_1=\cdot \cdot \cdot =D_N$. The BS and these devices collaborate to train an ML model by exchanging the models and gradients without sharing these locally stored raw data, which offers basic protection for users’ personal information. However, the BS is assumed to be curious and attempts to probe sensitive information from the received gradients, threatening users' privacy. In this work, the privacy of the scheduled devices can be guaranteed by channel noise by designing the alignment coefficient.

The goal of an FL task is to obtain the optimal model parameterized by $\boldsymbol{m}^*$ by minimizing the average global loss $L\left( \boldsymbol{m} \right)$, i.e.,
	\begin{equation}\label{proj5-eq1}
		\begin{aligned}
			\boldsymbol{m}^*= \arg\min_{\boldsymbol{m}} L\left( \boldsymbol{m} \right)\triangleq\frac{1}{N}\sum_{k=1}^N{L_k\left( \boldsymbol{m} \right)}
		\end{aligned}
	\end{equation}
	where $\boldsymbol{m }\in \mathbb{R} ^d$ is the model parameter to be optimized. More specifically, the objective function of device $k$ is defined as:
	\begin{equation}\label{proj5-eq2}
		\begin{aligned}
			L_k\left( \boldsymbol{m} \right) =\frac{1}{D_k}\sum_{\left( \boldsymbol{u},v \right) \in \mathcal{D} _k}{l\left( \boldsymbol{m};\left( \boldsymbol{u},v \right) \right)},
		\end{aligned}
	\end{equation}
	where $l\left( \boldsymbol{m};\left( \boldsymbol{u},v \right) \right)$ denotes the loss function, quantifying the error of model $\boldsymbol{m}$ on the input-output data pair $\left( \boldsymbol{u},v \right)$. 
\subsection{Over-the-Air Federated Averaging}
To solve the problem in (1) while reducing the communication overhead, we employ the classic and widely-adopted federated averaging (FedAvg) algorithm, which is implemented in an iterative manner. 
Generally, it requires a number of global aggregations, i.e., communication rounds, between devices and the BS to achieve the desired accuracy level of the learned global model $\boldsymbol{m}$. 
Specifically, we assume that $T$ and $I$ are the number of total training rounds and the number of communication rounds, respectively. 
Consequently, the local training step in each communication round is decided by $E = \frac{T}{I}$, and we assume that $T$ is divisible by $I$ \footnote{Since $I$ and $E$ are in one-to-one correspondences when we have a fixed $T$, we use $E$ and $I$ exchangeably when we discuss the impact of the communication rounds $I$ in the rest of this paper.}. 
Specifically, in each communication round $i \in \left \{ 0,\cdots,I-1 \right \}$, FedAvg consists of the following steps:
\emph {(1) Parameter broadcasting}:  At the beginning of communication round $i$, the BS broadcasts the latest global model parameter $\boldsymbol{m}^{i}$ to the scheduled devices denoted by $\mathcal{K}$, $\mathcal{K} \in \mathcal{N}$.
\emph{(2) Local training}: Each device first performs the initialization of the local model by setting the received global model parameter as the initial local model parameter, i.e., $\boldsymbol{w}_{k}^{i,0} = \boldsymbol{m}^{i}, \forall k \in \mathcal{K}$. Then,  each device performs $E$ rounds of local training by 
	\begin{align}\label{proj5-eq3}
		\boldsymbol{w}_{k}^{i,\iota+1} =\boldsymbol{w}_{k}^{i,\iota }-\tau  \nabla L_k\left( \boldsymbol{w}_{k}^{i, \iota} \right) , \iota \in \left \{ 0,...,E-1	 \right \},
	\end{align}
 where $\tau$ is the learning rate and 
	\begin{equation}\label{proj5-eq4}
		\begin{aligned}
			\nabla L_k\left( \boldsymbol{w}_{k}^{i,\iota } \right) =\frac{1}{D_k}\sum_{\left( \boldsymbol{u},v \right) \in \mathcal{D} _k}{\nabla l\left( \boldsymbol{w}_{k}^{i,\iota };\left( \boldsymbol{u},v \right) \right)}.
		\end{aligned}
	\end{equation}
\emph{(3) Over-the-air aggregation}: Upon completing $E$ times of local training, each scheduled device uploads the accumulative gradients in this current communication round to the BS, i.e., 
	\begin{equation}\label{proj5-eq5}
		\begin{aligned}
			\boldsymbol{g}_{k}^{i} & = \frac{1}{\tau}\left (\boldsymbol{w}_{k}^{i,E}-\boldsymbol{w}_{k}^{i,0} \right ) = \sum_{\iota  = 0}^{E-1} \nabla L_k\left( \boldsymbol{w}_{k}^{i,\iota } \right).
		\end{aligned}
	\end{equation}
To further alleviate communication bottlenecks and unbearable upload latency, we adopt analog over-the-air aggregation in this work, which enables the scheduled devices to simultaneously communicate their gradients to the BS via a shared MAC. Taking device $k$ as an example, the gradient is transmitted by a pre-processed signal $\boldsymbol{x} _{k}^{i}$:
	\begin{equation}\label{proj5-eq6}
		\begin{aligned}
			\boldsymbol{x} _{k}^{i}=e^{-j\psi _{k} }\left ( \frac{\sqrt{\varphi _kP_k} }{\varpi } \boldsymbol{g} _{k}^{i} \right ),
		\end{aligned}
	\end{equation}
where $e^{-j\psi _k}$ is the local phase correction performed by the device $k$. $P_k$ is the maximum transmission power of device $k$ and $\varphi _k\in \left[ 0,1 \right] $ is the power scaling factor. We assume that the upper bound of each gradient’s 2-norm is $\varpi$, i.e., $\left\| \boldsymbol{g}_{k}^{i} \right\| _2 \le \varpi$, so that $\mathrm {E}   \left [ \left \|  \boldsymbol{x} \right \| _{2}^{2}  \right ] \le P_{k} $. 
The scheduled devices upload their local gradients $\boldsymbol{g}_{k}^{i}$ via the uncoded form with perfect time synchronization among them. In this way, the gradients can be aggregated over the air thanks to the superposition property of MAC. Consequently, the received signal at the BS is given by
	\begin{equation}\label{proj5-eq7}
		\begin{aligned}
			\boldsymbol{y}^i=&\sum_{k\in \mathcal{K}}{h_k\boldsymbol{x}_{k}^{i}}+\boldsymbol{r}^i
			\\
			=&\sum_{k\in \mathcal{K}}{\left| h_k \right|\frac{\sqrt{\varphi _kP_k}}{\varpi}\boldsymbol{g}_{k}^{i}}+\boldsymbol{r}^i,
		\end{aligned}
	\end{equation} 
where $h_k=\left| h_k \right|e^{j\psi _k}$ is the complex-valued time-invariant channel coefficient between device $k$ and the BS. The received noise $\boldsymbol{r}^i\sim \mathcal{N} \left( 0,\sigma ^2\mathbf{I}_d \right) $ at the BS is employed to prevent privacy leakage in this paper. To recover the desired average gradient $\boldsymbol{g}^i=\frac{1}{\left| \mathcal{K} \right|}\sum_{k\in \mathcal{K}}{\boldsymbol{g}_{k}^{i}}$ from the received signal, the BS performs the post-processing by 
		\begin{equation}\label{proj5-eq8}
			\begin{aligned}
				\boldsymbol{\tilde{g}}^i=\frac{1}{\left| \mathcal{K} \right|\nu}\boldsymbol{y}^i
				=\frac{1}{\left| \mathcal{K} \right|}\sum_{k\in \mathcal{K}}{\left| h_k \right|\frac{\sqrt{\varphi _kP_k}}{\nu\varpi}\boldsymbol{g}_{k}^{i}}+\frac{1}{\left| \mathcal{K} \right|\nu }\boldsymbol{r}^i, 
			\end{aligned}
		\end{equation}
where $\nu$ is a post-processing factor, which is referred to as the alignment coefficient. The induced error between the recovered gradient and the desired gradient is derived as,
	\begin{equation}\label{proj5-eq9}
		\begin{aligned}
			\Delta \boldsymbol{g}^i_{err}=\underset{fading\ error}{\underbrace{\frac{1}{\left| \mathcal{K} \right|}\sum_{k\in \mathcal{K}}\left ( \left| h_k \right|\frac{\sqrt{\varphi _kP_k}}{\nu\varpi}-1 \right )\boldsymbol{g}_{k}^{i}}}+\underset{noise \ error}{\underbrace{\frac{1}{\left| \mathcal{K} \right|\nu }\boldsymbol{r}^i}}.
		\end{aligned}
	\end{equation}
The estimate gradient recovered from the over-the-air aggregated gradient results in two sources of error, i.e., the misalignment error due to fading and the additive error due to the noise.
In order to eliminate the fading-related error, the gradients need to be aligned by the alignment coefficient $\nu$ by adjusting the power scaling factor $\varphi _k$ in pre-precessing as follows,
	\begin{equation}\label{proj5-eq10}
		\begin{aligned}
			\left| h_k \right|\frac{\sqrt{\varphi _kP_k}}{\varpi}=\nu,\forall k\in \mathcal{K}, 
		\end{aligned}
	\end{equation}
which is referred to as the aligned OTA-FL and was also studied in \cite{seif2020wireless}. Following such an aligned aggregation scheme,  the received signal at the BS  in (\ref{proj5-eq7}) can be simplified as:
	\begin{equation}\label{proj5-eq11}
		\begin{aligned}                                  
			\boldsymbol{y}^i =\nu \sum_{k\in \mathcal{K}}{\boldsymbol{g}_{k}^{i}}+\boldsymbol{r}^i,
		\end{aligned}
	\end{equation}
and the estimated average gradient is finally given by,
	\begin{equation}\label{proj5-eq12}
		\begin{aligned}
			\boldsymbol{\tilde{g}}^i
			=\frac{1}{\left| \mathcal{K} \right|}\sum_{k\in \mathcal{K}}{\boldsymbol{g}_{k}^{i}}+\frac{1}{\left| \mathcal{K} \right|\nu }\boldsymbol{r}^i.
		\end{aligned}
	\end{equation}
\emph{(4) Model update}: The BS updates the global model parameter based on the estimated average gradient as follows:
	\begin{equation}\label{proj5-eq13}
		\begin{aligned}
			\boldsymbol{m}^{i+1}=\boldsymbol{m}^{i}-\tau \boldsymbol{\tilde{g}}^i.
		\end{aligned}
	\end{equation}
The above iteration steps are repeated until a certain training termination condition is met.
	
\subsection{Power Constraints of the OTA-FedAvg System}
In this paper, we consider both the peak transmit power constraint of each device and the sum power constraint of the overall DP-OTA-FedAvg system.
\subsubsection{Peak power constraint}
Following (\ref{proj5-eq10}), we have
	\begin{equation}\label{proj5-eq14}
		\begin{aligned}
			\varphi _k=\frac{\nu ^2\varpi ^2}{\left| h_k \right|^2P_k},\forall k\in \mathcal{K} .
		\end{aligned}
	\end{equation}
To make sure that $\varphi _k\le 1$, the alignment coefficient $\nu$ needs to satisfy:
	\begin{equation}\label{proj5-eq15}
		\begin{aligned}
			\nu \le \frac{\underset{s\in \mathcal{K}}{\min}\left\{ \left| h_s \right|\sqrt{P_s} \right\}}{\varpi}.
		\end{aligned}
	\end{equation}
From (\ref{proj5-eq15}), we can learn that the alignment coefficient $\nu$ is limited by the device with the worst channel condition among the scheduled devices, i.e., $\underset{s\in \mathcal{K}}{\min}\left\{ \left| h_s \right|\sqrt{P_s} \right\}$.
However, a larger $\nu$ is expected to mitigate the noise-related error following (\ref{proj5-eq9}). 
Since the learning performance will be degraded due to a small $\nu$, i.e., large noise error, the optimal design of device scheduling to improve the alignment coefficient is significant, especially in the FL systems where devices are power-limited and some of the devices suffer from poor channel conditions.
\subsubsection{Sum power constraint}
In each communication round, the power consumption for transmitting gradient of device $k$ is:
	\begin{equation}\label{proj5-eq16}
		\begin{aligned}
			\varphi _kP_k=\frac{\nu ^2\varpi ^2}{\left| h_k \right|^2},\forall k\in \mathcal{K}.
		\end{aligned}
	\end{equation}
Assume that the total power budget for the communication rounds of DP-OTA-FedAvg is $P^{tot}$. Then, the sum power transmit power constraint is given by,
	\begin{equation}\label{proj5-eq17}
		\begin{aligned}
			{\textstyle \sum_{k \in \mathcal{K} }} \frac{\nu ^2\varpi ^2}{\left| h_k \right|^2} \le \frac{P^{tot}}{I}.
		\end{aligned}
	\end{equation}
From (\ref{proj5-eq17}), we can learn that if the number of the communication rounds $I$ is small, the power budget in each communication round for gradient uploading will be large, which means that we can schedule more devices or set a large alignment coefficient. The impact of the number of the communication rounds $I$, the number of the scheduled devices, and the alignment coefficient $\nu$ on the learning performance will be discussed in Section \ref{proj5-section3}.

\subsection{Differential Privacy}\label{differentialprivacy}
DP \cite{dwork2014algorithmic} is defined on the conception of the adjacent dataset, which guarantees the probability that any two adjacent datasets output the same result is less than a constant with the help of adding random noise. More specifically, DP quantifies information leakage in FL by measuring the sensitivity of the gradients to the change of a single data point in the input dataset. The basic definition of $\left( \epsilon,\xi  \right)$-DP is given as follows.
\newtheorem{definition}{Definition}
	\begin{definition}$\left( \epsilon ,\xi \right)$-DP \cite{dwork2014algorithmic}:
		A randomized mechanism $\mathcal{O}$ guarantees $\left( \epsilon ,\xi  \right)$-DP if for two adjacent datasets $\mathcal{D} ,\mathcal{D} '$ differing in one sample, and measurable output space $\mathcal{Q} $ of $\mathcal{O}$, it satisfies,
		\begin{equation}\label{wg11}
			\begin{aligned}
				\mathrm{Pr}\left[ \mathcal{O} \left( \mathcal{D} \right) \in \mathcal{Q} \right] \le e^{\epsilon}\mathrm{Pr}\left[ \mathcal{O} \left( \mathcal{D} ' \right) \in \mathcal{Q} \right] +\xi .
			\end{aligned}
		\end{equation} 
	\end{definition}
	The additive term $\xi $ allows for breaching $\epsilon$-DP with the probability $\xi $ while $\epsilon$ denotes the protection level and a smaller $\epsilon$ means a higher privacy preservation level. Specifically, the Gaussian DP mechanism which guarantees privacy by adding artificial Gaussian noise is introduced as follows.
	\begin{definition}Gaussian mechanism \cite{dwork2014algorithmic}:
		A mechanism $\mathcal{O}$ is called as a Gaussian mechanism, which alters the output of another algorithm $\mathcal{L} :\mathcal{D} \rightarrow \mathcal{Q}$ by adding Gaussian noise, i.e., 
		\begin{equation}\label{wg11}
			\begin{aligned}
				\mathcal{O} \left( \mathcal{D} \right) =\mathcal{L} \left( \mathcal{D} \right) +\mathcal{N} \left( 0,\sigma ^2\mathbf{I}_d \right).
			\end{aligned}
		\end{equation} 
		Gaussian mechanism $\mathcal{O}$ guarantees $\left(\epsilon ,\xi  \right)$-DP with $\epsilon =\frac{\varDelta S}{\sigma}\sqrt{2\ln \left( \frac{1.25}{\xi } \right)}$ where $\varDelta S\triangleq \underset{\mathcal{D} ,\mathcal{D} '}{\max}\left\| \mathcal{L} \left( \mathcal{D} \right) -\mathcal{L} \left( \mathcal{D} ' \right) \right\| _2$ is the sensitivity of the algorithm $\mathcal{L}$  quantifying the sensitivity of the algorithm $\mathcal{L}$ to the change of a single data point. 
	\end{definition}
	
	\section{Privacy and Convergence Analysis of DP-OTA-FedAvg}\label{proj5-section3}
To reveal the impact of over-the-air aggregation on privacy and learning performance, we conduct privacy and convergence analysis in this section. Then, based on these analytical results, we formulate an optimization problem to minimize the optimality gap by optimizing the device scheduling, alignment coefficient, and the number of communication rounds subject to privacy and sum power constraints.

For analysis purposes, we provide the following common assumptions first.
\begin{assumption}\label{proj5-ass1}
		The expected squared norm of each gradient is bounded:
		\begin{equation}\label{proj5-eq20}	
			\begin{aligned}
				\mathbb{E} \left[ \left\| \boldsymbol{g}_{k}^{i} \right\| _2 \right] \le \varpi .
			\end{aligned}
		\end{equation}
	\end{assumption}
	
	\begin{assumption}\label{proj5-ass2}
		Assume that $L\left( \cdot \right) \,\,$ is $\zeta$-smooth, i.e., for all $\boldsymbol{\iota }'$ and $\boldsymbol{\iota }$, one has
		\begin{equation}\label{wg8}
			\begin{aligned}
				L\left( \boldsymbol{\iota }' \right) -L\left( \boldsymbol{\iota } \right) \le \left( \boldsymbol{\iota }'-\boldsymbol{\iota } \right) ^{\mathrm{T}}\nabla L\left( \boldsymbol{\iota } \right) +\frac{\zeta}{2}\left\| \boldsymbol{\iota }'-\boldsymbol{\iota } \right\| _{2}^{2}.
			\end{aligned}
		\end{equation}
	\end{assumption}
	
	\subsection{Privacy Analysis}
We aim to improve the learning performance while achieving a certain level of DP of the participants in the OTA-FedAvg system by designing device scheduling and alignment coefficient. We conduct the privacy analysis based on the Gaussian mechanism of DP in the following. To calculate the privacy leakage according to the Gaussian mechanism, the key point is the sensitivity of the OTA- FedAvg algorithm to the change of a single data point in the input dataset.
Taking device $m$ as an example, assume that $\mathcal{D} _{m}$ and $\mathcal{D} _{m}^{'}$ are two adjacent datasets differing in one sample, and $\boldsymbol{g}_{m}^{i} $ and ${\left( \boldsymbol{g}_{m}^{i} \right) '}$ are the two gradients obtained based on $\mathcal{D} _{m}$ and $\mathcal{D} _{m}^{'}$, respectively.
The two signals received at the BS corresponding to datasets $\mathcal{D} _m$ and $\mathcal{D} _{m}^{'}$ are given by
		\begin{equation}\label{proj5-eq22}
		\begin{aligned}
			\boldsymbol{y}^i &=\nu \sum_{k\in \mathcal{K}}{\boldsymbol{g}_{k}^{i}}+\boldsymbol{r}^i,
			\\
			\left( \boldsymbol{y}^i \right) '&=\nu \sum_{k\in \mathcal{K},k\ne m}\boldsymbol{g}_{k}^{i}+{\left( \boldsymbol{g}_{m}^{i} \right) '}+\boldsymbol{r}^i,
		\end{aligned}
	\end{equation} 
which only differ in the gradient from device $m$.
Then, the sensitivity of the OTA-FedAvg is given by $\varDelta S_{m}^{i}\triangleq \underset{\mathcal{D} _{m},\mathcal{D} _{m}^{'}}{\max}\left\| \boldsymbol{y}^i-\left( \boldsymbol{y}^i \right) ' \right\| _2$ and we have the following results.
	\begin{lemma}\label{proj5-lemma1}
		Assume that Assumption 1 holds and the set of the scheduled devices is $\mathcal{K}$. For each device $k\in\mathcal{K}$, such a OTA-FedAvg  algorithm achieves  $\left( \epsilon _k,\xi  \right)$-DP in each communication round where
		\begin{equation}\label{proj5-eq23}
			\begin{aligned}
				\epsilon _k=\frac{2\varpi \nu }{\sigma}\cdot \sqrt{2\ln \frac{1.25}{\xi }},k\in\mathcal{K}.
			\end{aligned}
		\end{equation}
	\end{lemma} 
	\textit{Proof:} Accroding to the definiation of sentivisity and (\ref{proj5-eq22}), we have
	\begin{equation}\label{proj5-eq24}
		\begin{aligned}
			\varDelta S_{m}^{i}\triangleq& \underset{\mathcal{D} _{m},\mathcal{D} _{m}^{'}}{\max}\left\| \boldsymbol{y}^i-\left( \boldsymbol{y}^i \right) ' \right\| _2=\nu\underset{\mathcal{D} _{m},\mathcal{D} _{m}^{'}}{\max}\left\|  \boldsymbol{g}_{m}^{i}-\left( \boldsymbol{g}_{m}^{i} \right) ' \right\| _2
			\\
			=&\nu\left\| \boldsymbol{g}_{m}^{i}-\left( \boldsymbol{g}_{m}^{i} \right) ' \right\| _2\overset{\left( a \right)}{\le}2\varpi\nu,
		\end{aligned}
	\end{equation}
	where (a) is from triangular inequality and Assumption \ref{proj5-ass1}.
	In accordance with the Gaussian mechanism of DP and replacing $m$ with $k$, one completes the proof of Lemma \ref{proj5-lemma1}.
	$\hfill \blacksquare$ 
	
Lemma \ref{proj5-lemma1} characterizes the impact of the alignment coefficient on privacy protection. More specifically, a smaller alignment coefficient $\nu$ leads to less privacy leakage. Physically speaking, a smaller alignment coefficient $\nu$ decreases the amplitude of the gradient signal, which enables the gradient more easily hidden in the channel noise. However, it degrades the utility of the gradients, which is validated in the following convergence analysis results.
	\begin{remark}
		Note that when the ``$=$" in (\ref{proj5-eq23}) is replaced by ``$\le$", it indicates a stronger privacy protection so it still satisfies $\left( \epsilon _{k},\xi  \right)$-DP.
	\end{remark}
\subsection{Convergence Analysis}
We here present convergence analysis in the cases of convex and non-convex loss functions. We first present the expectation of the gap between the updated global model $ \boldsymbol{m}^{i+1} $ and the current global model $ \boldsymbol{m}^{i} $ for the following analysis.
	\begin{lemma}\label{proj5-lemma2}
		Given the learning rate $\tau \le\frac{1}{\zeta}$, the upper bound of the gap between the updated global model $\boldsymbol{m}^{i+1}$ and the current model $\boldsymbol{m}^{i}$, i.e., $\mathbb{E} \left[ L\left( \boldsymbol{m}^{i+1} \right) \right] - \mathbb{E} \left[ L\left( \boldsymbol{m}^{i} \right) \right] 	$ is given by
		\begin{equation}\label{proj5-eq25}
			\begin{aligned}
				&\mathbb{E} \left[ L\left( \boldsymbol{m}^{i+1} \right) \right] - \mathbb{E} \left[ L\left( \boldsymbol{m}^{i} \right) \right] 	
				\\
				\le&-\frac{\tau}{2}\mathbb{E} \left[ \left\| \nabla L\left( \boldsymbol{m}^i \right) \right\| _{2}^{2} \right] 
				+\tau \varpi ^2\left( E-1 \right) ^2 +4\tau\varpi ^2\left( 1-\frac{\left| \mathcal{K} \right|}{N} \right) ^2
					+\frac{\zeta \tau ^2}{2}\frac{d\sigma ^2}{\left| \mathcal{K} \right|^2\nu ^2}.
				\end{aligned}
		\end{equation}
	The expectation is with respect to the randomness of Gaussian noise.	
	\end{lemma} 
	\textit{Proof:} Please refer to Appendix \ref{proj5-lemma2-proof}.$\hfill \blacksquare$ 

For notation simplicity, we define $\theta = \nu\varpi$ as an equivalent substitution of $\nu$ and refer to it as the alignment factor. In the following, we mainly focus on the alignment factor $\theta$ instead of $\nu$. Based on Lemma \ref{proj5-lemma2}, we give the following convergence analysis in both convex and non-convex settings.
\subsubsection{Convex Setting}
	We first consider the most benign setting, where the loss function $L\left( \cdot  \right)$ is assumed to be strongly convex. We formalize a strong convexity assumption as below.
	\begin{assumption}\label{proj5-ass3}
		Assume that $L\left( \cdot \right) \,\,$ is strongly convex with a positive parameter $\varrho$, i.e., for all $\boldsymbol{\iota }'$ and $\boldsymbol{\iota }$, one has
		\begin{equation}\label{proj5-eq26}
			\begin{aligned}
				L\left( \boldsymbol{\iota }' \right) -L\left( \boldsymbol{\iota } \right) \geqslant \left( \boldsymbol{\iota }'-\boldsymbol{\iota } \right) ^{\mathrm{T}}\nabla L\left( \boldsymbol{\iota } \right) +\frac{\varrho}{2}\left\| \boldsymbol{\iota }'-\boldsymbol{\iota } \right\| _{2}^{2}.
			\end{aligned}
		\end{equation}
	\end{assumption}
Under Assumption 3, we could derive a useful result \cite{bottou2018optimization} as follows:
	\begin{equation}\label{proj5-eq27}
		\begin{aligned}
			\left\| \nabla L\left( \boldsymbol{\iota } \right) \right\| _{2}^{2}\geqslant 2\varrho\left[L\left( \boldsymbol{\iota } \right) -L\left( \boldsymbol{\iota }^* \right) \right].
		\end{aligned}
	\end{equation}
We state the convergence theorem of the DP-OTA-FedAvg, describing its behavior when minimizing a strongly convex objective function with a fixed learning rate in the following.
	\begin{theorem}\label{proj5-theorem1}
		Assume that $\boldsymbol{m}^* $ is the optimal model and $\boldsymbol{m}^I$ is the obtained model after $I$ communication rounds. Assume that the learning rate is $\tau =\frac{1}{\zeta}$, then, the upper bound of the optimality gap $\mathbb{E} \left[ L\left( \boldsymbol{m}^I \right) -L\left( \boldsymbol{m}^* \right) \right]$ is given by
		\begin{equation}\label{proj5-eq28}
			\begin{aligned}
				&\mathbb{E} \left[ L\left( \boldsymbol{m}^I \right) -L\left( \boldsymbol{m}^* \right) \right] 
				\\
				\le& \eta ^{I}\underset{Initial\ gap}{\underbrace{\mathbb{E} \left[ L\left( \boldsymbol{m}^0 \right) -L\left( \boldsymbol{m}^* \right) \right] }}+\frac{\varpi ^2}{\varrho} \left(1-\eta ^{I}\right) \left[\underset{\mathcal{A}}{\underbrace{4\left( 1-\frac{\left| \mathcal{K} \right|}{N} \right) ^2}} +\underset{\mathcal{B}}{\underbrace{\left( E-1 \right) ^2}}+\underset{\mathcal{C}}{\underbrace{\frac{1}{2}\frac{d\sigma ^2}{\left| \mathcal{K} \right|^2\theta ^2}}}\right],
			\end{aligned}
		\end{equation}
		where 
		\begin{equation}\label{convercoeff}
			\begin{aligned}
				\eta =1-\frac{\varrho}{\zeta}.
			\end{aligned}
		\end{equation} 
		The expectation is with respect to the randomness of Gaussian noise.
	\end{theorem} 
	\textit{Proof:} Please refer to Appendix \ref{proj5-theorem1-proof}.$\hfill \blacksquare$ 
	
The optimality gap presented in the right-hand side (RHS) of (\ref{proj5-eq28}) demonstrates the impact of device scheduling $\mathcal{K}$, alignment factor $\theta$, and the local training times $E$ on the learning process. Specifically, term $\mathcal{A}$ is the error caused by partial device participation. 
A larger $\left| \mathcal{K} \right|$ contributes to a smaller optimality gap, i.e., a better learning performance.
This can be understood that the channel noise leads to a smaller distortion to the gradient average when more devices are involved.
This term decreases as the number of the scheduled devices increases and will be eliminated with full device participation, i.e., $\left| \mathcal{K} \right|=N$. 
The local update error shown in term $\mathcal{B}$ increases with the number of local training times $E$. If $E=1$, i.e., the FedAvg becomes the conventional FL algorithm, this term goes to 0. 
Term $\mathcal{C}$ is the error caused by the channel noise, which can be controlled by designing the device scheduling and the alignment coefficient. From this term, we can learn that a larger number of participants and the alignment coefficient contribute to a smaller noise-related error. 

Furthermore, Theorem \ref{proj5-theorem1} offers the following important insights: 
(1) The impact of the number of the communication round $I$: Given a fixed $\eta$, the first term decreases with the number of communication round $I$ due to the fact that $\eta \le 1$. When $I$ goes to infinity, the first term approaches zero. 
For the second term, on one hand, a larger $I$ leads to a smaller $E$, which is beneficial for mitigating the local training error. On the other hand, a larger $I$ increases the weight of the design-related error, i.e., term $\mathcal{A}$,$\mathcal{B}$,$\mathcal{C}$.
(2) The impact of $\eta$: Given the fixed communication round $I$, a larger $\eta$ (closer to $1$) leads to a larger first term while a smaller second term.

Based on Theorem 1, we can also derive the optimality gap of a conventional FL algorithm where the number of local training times is one with full device participation and a noise-free channel.
	\begin{corollary} \label{corollaryoptimalitygap}
		Given the total training number is $T$ and learning rate $\tau =\frac{1}{\zeta}$, the upper bound of the optimality gap $\mathbb{E} \left[ L\left(\boldsymbol{m}^T \right) -L\left( \boldsymbol{m}^* \right) \right]$ of an conventional FL algorithm with one local training round without considering noise and device scheduling is
		\begin{equation}\label{proj5-eq30}
			\begin{aligned}
				\mathbb{E} \left[ L\left( \boldsymbol{m}^T \right) -L\left( \boldsymbol{m}^* \right) \right] 
				\le\left( 1-\frac{\varrho}{\zeta} \right) ^{T}\mathbb{E} \left[ L\left( \boldsymbol{m}^0 \right) -L\left( \boldsymbol{m}^* \right) \right] .
			\end{aligned}
		\end{equation}
	\end{corollary}
	\itshape {Proof:}  \upshape  If the FL algorithms with full device participation only perform one local training round in each communication round and communicate through the noiseless channel, we have $E=1$, $\left| \mathcal{K} \right|=N$, and $\sigma =0$. Hence, $4\left( 1-\frac{\left| \mathcal{K} \right|}{N} \right) ^2 +\left( E-1 \right) ^2 +\frac{1}{2}\frac{d\sigma ^2}{\left| \mathcal{K} \right|^2\theta ^2}=0$. Then (\ref{proj5-eq30}) can be derived based on (\ref{proj5-eq28}). 	
	\hfill $\blacksquare$
	
From Corollary \ref{corollaryoptimalitygap}, we can observe that, if the communication between the BS and devices is noise-free and there is only one local training in each communication round, the FL algorithm with all device participation will converge to the optimal global FL model without any gaps. This result corresponds to the results in the existing works \cite{chen2020joint,friedlander2012hybrid}. 
	
\subsubsection{Non-Convex Setting}
Considering that many useful machine learning models, e.g., deep neural networks, lead to non-convex objective functions, we thus investigate the convergence property of DP-OTA-FedAvg in the non-convex setting in the following. Different from the convex case where the expected optimality gap is employed to measure the convergence rate. In the case of non-convex loss function $L\left( \cdot \right) $, the algorithm converging to a global minimum cannot in general be guaranteed. A reasonable substitute is to study the convergence to local minimum, or at the very least, to stationary points \cite{ghadimi2013stochastic,drori2020complexity}. Hence, similar to previous work \cite{sun2021pain,zhang2021adaptive,liu2022chronos,chen2022energy}, we use the relationship between the average expected squared gradient norm and the iteration number to characterize the convergence rate of DP-OTA-FedAvg.
\begin{theorem}\label{proj5-theorem2}
	Given the learning rate $\tau$ and the communication rounds $I$, the average-squared gradient after $I$ communication rounds is bounded as follows,	
		\begin{equation}\label{proj5-eq33}
			\begin{aligned}
				&\frac{1}{I}\sum_{i=0}^{I-1}{\mathbb{E} \left[ \left\| \nabla L\left( \boldsymbol{m}^i \right) \right\| _{2}^{2} \right]}
				\\
				\le &\frac{2}{\tau I}\left[ \mathbb{E} \left[ L\left( \boldsymbol{m}^0 \right) \right] -\left[ L\left( \boldsymbol{m}^* \right) \right] \right] +\varpi ^2\left[ 8\left( 1-\frac{\left| \mathcal{K} \right|}{N} \right) ^2+2\left( E-1 \right) ^2+\frac{d\sigma ^2}{\left| \mathcal{K} \right|^2\theta ^2} \right]. 
			\end{aligned}
		\end{equation}
		The expectation is with respect to the randomness of Gaussian noise.
	\end{theorem} 
	\textit{Proof:} Please refer to Appendix \ref{proj5-theorem2-proof}.$\hfill \blacksquare$ 

In Theorem \ref{proj5-theorem2}, we get an upper bound of the average-squared gradients of $L\left( \cdot \right) $ for a certain communication round number $I$. If the upper bound tends to 0, then the algorithm converges, because this implies that $\mathbb{E} \left[ \left\| \nabla L\left( \boldsymbol{m}^i \right) \right\| _{2}^{2} \right]\longrightarrow 0$. It can be found that the first term of the upper bound is inversely proportional to $I$, so it tends to 0 as $I$ approaches infinity. The second term can be reduced by designing the device scheduling, alignment factor, and the local training times.
	
\section{DP-OTA-FedAvg with limited sum power budgets}	\label{proj5-section4}
In order to improve the learning performance of DP-OTA-FedAvg with privacy and sum power constraints, we formulate the following problem where we take the optimality gap $\mathbb{E} \left[ L\left( \boldsymbol{m}^I \right) -L\left( \boldsymbol{m}^* \right) \right]$ as the objective function. To minimize the objective function, we expect a larger $\left|\mathcal{K}\right|$ and a larger $\theta$. However, $\theta$ is limited by the device in $\mathcal{K}$ with the worst channel condition. We can improve $\theta$ by scheduling the devices with better channel conditions to participate in the training, which leads to a smaller $\left|\mathcal{K}\right|$. Therefore, there is a tradeoff between the number of the scheduled devices $\left|\mathcal{K}\right|$ and the alignment factor $\theta$. On the other hand, the impact of the global aggregation $I$ is insignificant. A larger $I$, i.e., a smaller $E$ helps to mitigate the local update error and the initial gap, however, may introduce more transmission distortion. Therefore, the design of device scheduling, alignment factor, and the number of global aggregations is important for improving learning performance while preserving privacy.

	
	
\subsection{Problem Formulation}
Assume that each device has the same privacy budget $\left( \epsilon,\zeta  \right)$, i.e.,  the maximum value of tolerable privacy leakage. The total training rounds is $T$ and we use $\frac{T}{I}$ to substitute $E$ for simplicity. The number of global aggregations $I$ and local training times $E$ should be an integer. We firstly ignore the integer constraint of $E$, which will finally be guaranteed by rounding operation. By defining $G= \mathbb{E} \left[ L\left( \boldsymbol{m}^0 \right) \right] -\left[ L\left( \boldsymbol{m}^* \right) \right] $, $\phi = \sqrt{2\ln \frac{1.25}{\xi }}$, and the set of natural numbers $\mathcal{Z}$, the problem is formulated as follows.
	\begin{align}\label{proj5-eq34}
		\mathbf{P}1.&\quad \underset{I,\mathcal{K},\theta}{\min}\left\{ \eta ^{I}G+\frac{\varpi ^2}{\varrho} \left(1-\eta ^{I}\right) \left[4\left( 1-\frac{\left| \mathcal{K}\right|}{N} \right) ^2 +\left( \frac{T}{I}-1 \right) ^2 +\frac{1}{2}\frac{d\sigma ^2}{\left(\left| \mathcal{K} \right|\theta \right)^2}\right]\right\} 
			\\
			\mathbf{s}.\mathbf{t}.&\quad \mathcal{K} \subseteq \mathcal{N} ,
			\tag{\ref{proj5-eq34}a}
			\\
			& \quad \frac{2 \theta }{\sigma}\cdot \phi \le\epsilon, \tag{\ref{proj5-eq34}b}
			\\
			& \quad \theta \le \underset{s\in \mathcal{K}}{\min}\left\{ \left| h_s \right|\sqrt{P_s} \right\}, \tag{\ref{proj5-eq34}c}
			\\
			& \quad I  { \sum_{k\in \mathcal{K} }^{ }} {\frac{ \theta ^2}{\left | h_k \right |^2}}\le P^{tot}, \tag{\ref{proj5-eq34}d}
	    	\\
		  &\quad 1 \le I \le T,  I \in \mathcal{Z} \tag{\ref{proj5-eq34}e}.
		\end{align}
Constraint (\ref{proj5-eq34}a) guarantees that $\left|\mathcal{K}\right| \le N$; Constraint (\ref{proj5-eq34}b) ensures that the privacy leakage of each device does not exceed the privacy budget; Constraint (\ref{proj5-eq34}c) implies that the alignment coefficient should ensure that $\varphi _k\le 1$ as mentioned in (\ref{proj5-eq15}) due to the peak power constraint; Constraint (\ref{proj5-eq34}d) is the sum power constraint. Constraint (\ref{proj5-eq34}e) implies that the number of the aggregation rounds should be an integer and no more than $T$. 

P1 is solved in the following way. First, we decouple P1 into two sub-problems. Given the number of communication rounds, the set of potential optimal solution pairs is obtained by exploring the relationship between the number of scheduled devices and the alignment coefficient. Thanks to the reduced search space, the globally optimal device scheduling policy and alignment factor can be efficiently found by searching the set of the limited solution pairs. Futhermore, based on the optimal scheduling and alignment factor, the optimal communication times can be obtained by the one-dimensional search.
	
\subsection{Optimal Device Scheduling and Alignment Factor}
Assume that the optimal number of the global aggregation rounds is $I^*$ and define $c_{\left[ \mathcal{K} \right]}=\underset{s\in \mathcal{K}}{\min}\left\{ \left| h_s \right|\sqrt{P_s} \right\}$ and $q_{\left [ \mathcal{K}  \right ] } =\sqrt{\frac{P^{tot}}{I^*}}\left ( 1/\sqrt{\sum_{k\in \mathcal{K}}\left ( 1/\left | h_k \right |^2 \right ) } \right ) $ for notation simplicity. Then, the constraints (\ref{proj5-eq34}b), (\ref{proj5-eq34}c) and (\ref{proj5-eq34}d) can be rewritten as $\theta \le \min \left\{ \frac{\epsilon \sigma}{2\phi},  c_{\left[ \mathcal{K} \right]},q_{\left [ \mathcal{K}  \right ] }  \right\} $. The problem that optimizes device scheduling $\mathcal{K}$ and alignment factor $\theta$ can be decoupled as follows:
		\begin{align}\label{proj5-eq35}
			\mathbf{P}2.& \quad \underset{\mathcal{K} ,\theta}{\min}\left\{4\left( 1-\frac{\left| \mathcal{K} \right|}{N} \right) ^2+\frac{1}{2}\frac{d\sigma ^2}{\left| \mathcal{K} \right|^2\theta ^2}\right\} 
			\\
			\mathbf{s}.\mathbf{t}.&\quad \mathcal{K} \subseteq \mathcal{N}, 
			\tag{\ref{proj5-eq35}a}
			\\
			& \quad \theta \le \min \left\{ \frac{\epsilon \sigma}{2\phi}, c_{\left[ \mathcal{K} \right]},  q_{\left [ \mathcal{K}  \right ] }  \right\} .\tag{\ref{proj5-eq35}b}
		\end{align}
By observing the objective function, we know that larger $\left| \mathcal{K} \right|$ and $\theta$ yield a better objective function value. However, the upper bound of $\theta$ is limited by the scheduling policy $\mathcal{K}$ regarding the constraint (\ref{proj5-eq34}c). To this end, we first analyze the relationship between the number of scheduled devices $\left|\mathcal{K}\right|$ and the alignment factor $\theta$, which offers a set of potential optimal solution pairs. 
	
For clarity, we first consider the special case that all the devices are with the same transmit power budget $P^{dev}$, i.e., $P_1=P_2=...=P_N=P^{dev}$ and the general situation that each device with distinct peak power budget $P _k$ will be discussed in Subsection \ref{proj5-subsectionE}. Assume that the devices are sorted in ascending order of $\left | h_k\right |$, i.e., $\left | h_1\right |\le \left | h_2\right | \le ...\le \left | h_N\right |$. 
	In order to specify the relationship between $\left|\mathcal{K}\right|$ and $\theta$, we first conclude the relationship between $\left | \mathcal{K}  \right | $ and the upper bounds of $c_{\left[ \mathcal{K} \right]}$ and $q_{\left[ \mathcal{K} \right]}$, which limits the value of $\theta$.
	\begin{lemma}\label{proj5-lemma3}
		Assume that $c_{\left | \mathcal{K}  \right | }^{max}$and $q_{\left | \mathcal{K}  \right | }^{max}$ are the achieveable upper bounds of $c_{\left[ \mathcal{K} \right]}$ and $q_{\left[ \mathcal{K} \right]}$ for a given $\left | \mathcal{K}  \right | $,  which can be given as follows:
		\begin{equation}
			\begin{aligned}\label{proj5-eq36}
				c_{\left | \mathcal{K}  \right | }^{max}= \left | h_{N-\left | \mathcal{K} \right | +1} \right | \sqrt{P^{dev}}, \quad q_{\left | \mathcal{K}  \right | }^{max}=\sqrt{\frac{P^{tot}}{I^*}}\left ( 1/\sqrt{\sum_{j={N-\left | \mathcal{K} \right | +1}}^N\left ( 1/\left | h_j\right |^2 \right ) } \right ),
			\end{aligned}	 	
		\end{equation}
	in which case, the scheduling policy $\mathcal{K}$ is given by
	\begin{equation}
		\begin{aligned}\label{proj5-eq37}
			\mathcal{K} =\left \{ k\mid k\ge N-\left | \mathcal{K} \right | +1 \right \}.
		\end{aligned}	 	
	\end{equation}
	\end{lemma} 
	
\itshape {Proof:}  \upshape Assume that $\left | \mathcal{K}  \right | = j$, different $\mathcal{K}$ of size $j$ leads to different $c_{\left[ \mathcal{K} \right]}$ and $ q_{\left [ \mathcal{K}  \right ] }$. For example, if $\mathcal{K}_1=\left \{ k\mid  1 \le k \le j \right\}$, we have $c_{\left[ \mathcal{K}_1 \right]} =  \left | h_{1} \right | \sqrt{P^{dev}}$, while $\mathcal{K}_2=\left \{ k\mid  N-j+1 \le k \le N \right\}$, we have $c_{\left[ \mathcal{K}_2\right]} =  \left | h_{N-j+1} \right | \sqrt{P^{dev}}$. Since $\left | h_{N-j+1} \right | > \left | h_{1} \right |$, we have $c_{\left[ \mathcal{K}_2\right]}> c_{\left[ \mathcal{K}_1\right]}$.
Therefore, given a value of $\left|\mathcal{K}\right|$, the upper bound of $c_{\left[ \mathcal{K} \right]}$ can be obtained by scheduling the devices with better channel conditions.
Specifically, since $c_{\left[ \mathcal{K} \right]}$ is decided by the smallest $\left | h_k\right | , k\in \mathcal{K}$, the top-$\left | \mathcal{K}  \right |$ devices with the largest $\left | h_k\right | $ should be scheduled. Then, the upper bound of $c_{\left[ \mathcal{K} \right]}$ is given by $c_{\left[ \mathcal{K} \right]} = \left | h_{N-\left | \mathcal{K} \right | +1} \right | \sqrt{P^{dev}}$ where $\mathcal{K} =\left \{ k\mid k\ge N-\left | \mathcal{K} \right | +1 \right \}$.
Similarly, the larger $\left | h_k\right |$ contributes to a larger $q_{\left[ \mathcal{K} \right]}$ given a fixed value of $\left | \mathcal{K} \right |$. Therefore, the largest $q_{\left[ \mathcal{K}\right]}$ is obtained when $\mathcal{K} =\left \{ k\mid k\ge N-\left | \mathcal{K} \right | +1 \right \}$.
\hfill $\blacksquare$

Lemma \ref{proj5-lemma3} reveals an insight that if the $\left | \mathcal{K}  \right |$ is given, the upper bounds of $c_{\left[ \mathcal{K} \right]}$ and $q_{\left[ \mathcal{K} \right]}$ are determined, which corresponds to the same $\mathcal{K}$. In other words, the scheduling policy $\mathcal{K}$ that can achieve the largest value of alignment factor $\theta$ is obtained. More specifically, if the number of the scheduled device $\left | \mathcal{K}  \right |$ is given, the optimal scheduling policy $\mathcal{K}$ and alignment factor $\theta$ is obtained.
For example, if $\left | \mathcal{K}  \right |=1$, the optimal solution is given by $	\mathcal{K}^ *=\left \{N \right \}$ and $ \theta^* = \min \left\{ \frac{\epsilon \sigma}{2\phi}, c_{\left[ \mathcal{K}^* \right]},  q_{\left [ \mathcal{K}^* \right ] }  \right\}$. Since the optimal $\left | \mathcal{K}  \right |$ has not been obtained, each solution pair $\mathcal{K}^ *$ and $ \theta^*$ based on $\left | \mathcal{K}  \right |$ is the potential optimal solution. 
The maximum number of potential optimal solution pairs is $N$, i.e., the value of $\left | \mathcal{K} \right | $ is set from $1$ to $N$.  However, by further considering the potential value of $\theta$ constrained by $\theta \le \frac{\epsilon \sigma}{2\phi}$, we can narrow the space of the potential solution pairs as follows.
	
We define $\boldsymbol{c}=\left[ c_1,...c_m,...c_N \right]$ and $\boldsymbol{q}=\left[ q_1,...q_m,...q_N \right] $ where $c_m = \left | h_m \right | \sqrt{P^{dev}} $ and $q_m=\sqrt{\frac{P^{tot}}{I^*}}\left ( 1/\sqrt{\sum_{j=m}^N\left ( 1/\left | h_j\right |^2 \right ) } \right ) $. Since $\left | h_1\right |\le \left | h_2\right | \le ...\le \left | h_N\right |$, the elements in $\boldsymbol{c}$ and $\boldsymbol{q}$ are sorted in the ascending order.
Therefore, the minimal value of  $q_{\left [ \mathcal{K}  \right ] }$ and $c_{\left[ \mathcal{K} \right]} $ are  $q_1=\sqrt{\frac{P^{tot}}{I^*}}\left ( 1/\sqrt{\sum_{j=1}^N\left ( 1/\left | h_j\right |^2 \right ) } \right )$ and $c_1= \left | h_1 \right | \sqrt{P^{dev}} $, in which case $\mathcal{K}=\mathcal{N}$. For clarity, we give the solutions in two cases: 1) $ \frac{\epsilon \sigma}{2\phi}<\min \left\{ c_1,q_1 \right\} $; 2) $\min\left \{ c_1,q_1 \right \}  \le \frac{\epsilon \sigma}{2\phi}$ as follows.
	
	\subsubsection{In the case that $ \frac{\epsilon \sigma}{2\phi}<\min \left\{ c_1,q_1 \right\} $} The constraint of $\theta$ is independent of the device scheduling $\mathcal{K}$. Constraint (\ref{proj5-eq35}b) can be rewritten as $ \theta \le  \frac{\epsilon \sigma}{2\phi} $. Then, the optimal solution to $\mathbf{P}2$ can be given by the following Lemma.
	\begin{lemma}\label{proj5-lemma4}
		If $ \frac{\epsilon \sigma}{2\phi}<\min \left\{ c_1,q_1 \right\} $, the optimal solution to $\mathbf{P}2$ is
		\begin{align}\label{wg37}
			\theta ^* =\frac{\epsilon \sigma}{2\phi},\quad \mathcal{K} ^*=\mathcal{N},
		\end{align}	 	
		in which case all the devices are scheduled.
	\end{lemma} 
	\itshape {Proof:}  \upshape Firstly, to achieve a larger $\theta$, we have $\theta ^* =\frac{\epsilon \sigma}{2\phi}$. On the other hand, all the devices with $c_k\ge \frac{\epsilon \sigma}{2\phi}$ should be selected to achieve a larger $\left| \mathcal{K} \right|$, i.e., a better value of objective function.  Since $\frac{\epsilon \sigma}{2\phi}<  c_1\le c_k,\forall k\in \mathcal{N} $, we have $\mathcal{K} ^* =\mathcal{N}$. This completes the proof of Lemma \ref{proj5-lemma4}. 
	\hfill $\blacksquare$
	
	\subsubsection{In the case that $\min\left \{ c_1,q_1 \right \}  \le \frac{\epsilon \sigma}{2\phi}$}
	The constraint of $\theta$ is related to the device scheduling $\mathcal{K}$.
	 We assume that $\mathcal{Q} =\mathcal{Q}_1 \cup \mathcal{Q}_2$ where $\mathcal{Q}_1 =\left\{ \left. k \right|c_1\le c_k< \frac{\epsilon \sigma}{2\phi} \right\}$ and $\mathcal{Q}_2 =\left\{ \left. k \right|q_1\le q_k< \frac{\epsilon \sigma}{2\phi} \right\}$. 
	Then, we have $\min\left \{ c_{\left | \mathcal{Q}  \right | }, q_{\left | \mathcal{Q}  \right | } \right \} < \frac{\epsilon \sigma}{2\phi} \le \min\left \{ c_{\left | \mathcal{Q}  \right |+1 }, q_{\left | \mathcal{Q}  \right |+1 } \right \}$. Constraint (\ref{proj5-eq35}b) can be discussed in two cases: (1) $\theta \le \min\left \{ c_{\left [ \mathcal{K}  \right ] },q_{\left [ \mathcal{K}  \right ] } \right \}=\min\left \{ c_k,q_k \right \}  ,k\in\mathcal{Q} $; (2) $\theta \le \frac{\epsilon \sigma}{2\phi}$. Therefore, there are $\left | \mathcal{Q}  \right |+1$ potential upper bounds of $\theta$. For each upper bound of $\theta$, there is a corresponding optimal $\left | \mathcal{K}  \right | $ following Lemma \ref{proj5-lemma3}. Then, we have the following results.
	\begin{lemma}\label{proj5-lemma5}
		The minimum value of the potential optimal $\left| \mathcal{K} \right|$ is $N-\left| \mathcal{Q} \right|$. The relationship between the potential optimal solution pairs, i.e.,  $\left| \mathcal{K} \right|$ and $\theta$ can be given by
		\begin{align}\label{wg37}
			\theta=
			\begin{cases} \min \left\{c_{N-\left| \mathcal{K} \right|+1},q_{N-\left| \mathcal{K} \right|+1}\right\},if \left| \mathcal{K} \right| \ge N-\left| \mathcal{Q} \right|+1
				\\\frac{\epsilon \sigma}{2\phi}, if \left| \mathcal{K} \right| = N-\left| \mathcal{Q} \right|
			\end{cases}.
		\end{align}
	\end{lemma} 
	\itshape {Proof:}  \upshape 
	Firstly, given a value of $\left| \mathcal{K} \right| \ge N-\left| \mathcal{Q} \right|+1$, the largest $c_{\left[ \mathcal{K} \right] }$ and $q_{\left[ \mathcal{K} \right] }$ that can be achieved is $c_{N-\left| \mathcal{K} \right|+1}$ and $q_{N-\left| \mathcal{K} \right|+1}$. To achieve a larger $\theta$, we have $\theta=\min \left\{c_{N-\left| \mathcal{K} \right|+1},q_{N-\left| \mathcal{K} \right|+1}\right\}$. The largest feasible value of $\theta$ is $\frac{\epsilon \sigma}{2\phi}$, in which $\left| \mathcal{K} \right|$ achieves the minimum value $N-\left| \mathcal{Q} \right|$. Then, we complete the proof of Lemma \ref{proj5-lemma5}.
	\hfill $\blacksquare$
	
\begin{figure*}
		\centering
		\includegraphics[scale=1]{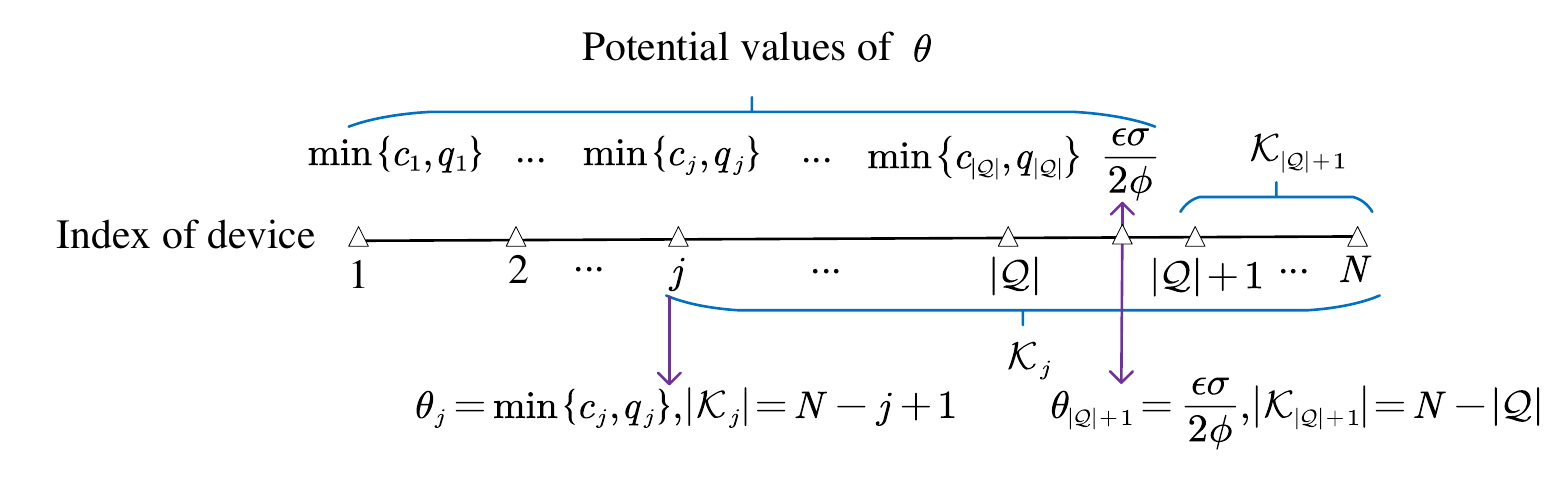}
		\caption{Illustration of the potential optimal solution space.}\label{proj5-fig2}
	\end{figure*}

	Then, the space of the potential optimal solutions pairs to $\mathbf{P}2$ as shown in Fig. \ref{proj5-fig2}, can be given as follows.
	\begin{lemma}\label{proj5-lemma6}
		There are $\left| \mathcal{Q} \right|+1$ closed-form solution pairs which may be the globally optimal solution. The $j$-th, $1\leq j \leq \left| \mathcal{Q} \right|$, solution pair $\theta_j$ and $\mathcal{K} _j$ is given by
		\begin{align}\label{wg37}
			\theta_j = \min\left \{ c_j,q_j \right \} , \quad \mathcal{K} _j=\left\{ \left. k \right|k\ge j \right\},
		\end{align}
		and the $\left| \mathcal{Q} \right|+1$-th solution pair $\theta_{\left| \mathcal{Q} \right|+1}$, $\mathcal{K} _{\left| \mathcal{Q} \right|+1}$ is 
		\begin{align}\label{wg37}
			 \theta_ {\left| \mathcal{Q} \right|+1}= \frac{\epsilon \sigma}{2\phi}, \quad \mathcal{K} _{\left| \mathcal{Q} \right|+1}=\left\{ \left. k \right|k \ge \left| \mathcal{Q} \right|+1\right\}.
		\end{align}
	\end{lemma}
	
	\itshape {Proof:}  \upshape Firstly, there are $\left| \mathcal{Q} \right|$ elements in $\mathcal{Q}$, which are the potential value of $ \min\left \{ c_{\left [ \mathcal{K}  \right ] },q_{\left [ \mathcal{K}  \right ] } \right \}$, i.e., $\theta$. It thus follows from Lemma \ref{proj5-lemma5} that there are $\left| \mathcal{Q} \right|$ pairs of $\theta$ and $\left| \mathcal{K} \right|$, i.e., $\left| \mathcal{Q} \right|$ potential optimal solution pairs, which may achieve the best performance. Specifically, the $j$-th solution corresponds to the setting that $\min\left \{ c_j,q_j \right \}$ and $\left| \mathcal{K} \right|=N-j+1$, in which case, $\mathcal{K} _j=\left\{ \left. k \right|k\ge j \right\}$. Additionally, $\theta =\frac{\epsilon \sigma}{2\phi}$ is the $\left| \mathcal{Q} \right|+1$-th solution, in which case we have $\left| \mathcal{K} \right|=N-\left| \mathcal{Q} \right|$ to achieve a better value of objective function and $\mathcal{K} _{\left| \mathcal{Q} \right|+1}=\left\{ \left. k \right|k \ge \left| \mathcal{Q} \right|+1\right\}$. Then, we complete the proof of Lemma \ref{proj5-lemma6}.
	\hfill $\blacksquare$
	
	Based on Lemma \ref{proj5-lemma6}, we can perform the one-dimension search method to obtain the optimal solution. The optimal solution to $\mathbf{P}2$ is $\mathcal{K} ^*, \theta^*$ where
	\begin{align}
		\mathcal{K} ^*, \theta^*=arg\underset{1\le j\le \left| \mathcal{Q} \right|+1}{\min}\left\{ \varPsi \left( \mathcal{K} _j,\theta _j \right) \right\} ,
	\end{align}
	where $\varPsi \left( \mathcal{K} _j,\theta _j\right) =4\left( 1-\frac{\left| \mathcal{K} _j \right|}{N} \right) ^2+\frac{d\sigma ^2}{2\left| \mathcal{K} _j\right|^2\theta _{j}^{2}}$. In fact, $\theta^*$ is the optimal threshold to schedule devices.
The overall procedure for solving $\mathbf{P}2$ is summarized in Algorithm 1.
	
From the above analysis, it can also be learned that the aligned DP-OTA-FL with device scheduling will not be worse than that with full devices participation  because the full device scheduling is one case of the potential optimal solution pairs. We next present which pairs of solutions for device scheduling can achieve better performance than the case of full device participation. Since the optimal solution is the same as full device situation when $\frac{\epsilon \sigma}{2\phi}<\min \left\{ c_1,q_1 \right\} $ as shown in Lemma \ref{proj5-lemma4}, we only consider the case that $\min\left \{ c_1,q_1 \right \}  \le \frac{\epsilon \sigma}{2\phi}$.
		
	\begin{algorithm}[ht]  
		\renewcommand{\algorithmicrequire}{\textbf{Input:}} 
		\renewcommand{\algorithmicensure}{\textbf{Output:}}
		\caption{The Procedure for  Solving Problem $\mathbf{P2 }$}  
		\label{Iteration}  
		\begin{algorithmic}[1]  
			\Require 
			Given $N$, $d$, $\sigma$, $\left ( \epsilon,\xi  \right ) $, $\boldsymbol{h}=\left \{ \left | h_1\right |,\cdot \cdot \cdot,  \left | h_N\right |\right \}$, $P ^{dev}$ and $P ^{tot}$. Initialize $I^*= T$.
			\Ensure  
			$\mathcal{K} ^*, \theta^*$.
			\State Calculate $ \frac{\epsilon \sigma}{2\phi}$, $\boldsymbol{c}$ and $\boldsymbol{q}$.
			\If {$ \frac{\epsilon \sigma}{2\phi}\le \min\left \{ c_1,q_1 \right \}  $ }
			\State  Obtain the optimal solution $\mathcal{K} ^*=\mathcal{N}, \theta ^* =\frac{\epsilon \sigma}{2\phi}$ following Lemma \ref{proj5-lemma4}.
			\Else 
			\State Calculate $\mathcal{Q} =\mathcal{Q}_1 \cup \mathcal{Q}_2$ where $\mathcal{Q}_1 =\left\{ \left. k \right|c_1\le c_k< \frac{\epsilon \sigma}{2\phi} \right\}$ and $\mathcal{Q}_2 =\left\{ \left. k \right|q_1\le q_k< \frac{\epsilon \sigma}{2\phi} \right\}$.
			\State Obtain the feasiable values of $\theta$ following Lemma \ref{proj5-lemma5}.
			\State Obtain $\left| \mathcal{Q} \right|+1$ pairs of potential optimal solution following Lemma \ref{proj5-lemma6}.
		    \State Obtain the optimal solutions by $\mathcal{K} ^*, \theta^*=arg\underset{1\le j\le \left| \mathcal{Q} \right|+1}{\min}\left\{ \varPsi \left( \mathcal{K} _j,\theta _j \right) \right\} $.
			\EndIf
		\end{algorithmic}  
	\end{algorithm}  

	\begin{lemma}\label{proj5-lemma7}
		If $\min\left \{ c_1,q_1 \right \}  \le \frac{\epsilon \sigma}{2\phi}$, the solution pairs $\mathcal{K}$ and $\theta$ that satisfies the following condition will make the aligned DP-OTA-FedAvg perform better than that with full device participation:
		\begin{align}\label{proj5-eq42}
			\left| \mathcal{K} \right|\theta\geqslant \frac{1}{\sqrt{\frac{1}{N^2c_{1}^{2}}-\frac{8}{d\sigma ^2}}}.
		\end{align}
	\end{lemma}
\itshape {Proof:}  \upshape The aligned DP-OTA-FedAvg with full device participation is equivalent to the solution that $\theta =  \min \left\{c_1,q_1\right\}$ and $\mathcal{K} = \mathcal{N}$, in which case, the value of the objective function is $\frac{d\sigma^2}{N^{2}c_1^2} $. 
By solving $4+\frac{d\sigma ^2}{2\left| \mathcal{K} \right|^2\theta ^2}\le \frac{d\sigma^2}{2N^{2}c_1^2} $, we complete the proof of Lemma \ref{proj5-lemma7}.
\hfill $\blacksquare$
	
	
\subsection{Optimal Number of Global Aggregation}
Given the optimal solutions of $\mathcal{K}$ and $\theta$ as $\mathcal{K^*}$ and $\theta ^*$, the problem of the optimal number of the global aggregations can be formulated by,
	\begin{align}\label{proj5-eq43}
		\mathbf{P}3.&\quad \underset{I}{\min}\left\{  \eta ^{I}G+\frac{\varpi ^2}{\varrho} \left(1-\eta ^{I}\right) \left[4\left( 1-\frac{\left| \mathcal{K}\right|}{N} \right) ^2 +\left( \frac{T}{I}-1 \right) ^2 +\frac{1}{2}\frac{d\sigma ^2}{\left(\left| \mathcal{K} \right|\theta \right)^2}\right]\right\} 
		\\
		\mathbf{s}.\mathbf{t}.& \quad 1 \le I \le \min\left\{ \frac{ P^{tot}}{{  \theta ^2\sum_{k\in \mathcal{K} }^{ }} {\frac{1}{\left | h_k \right |^2}}} ,T\right\},  \quad  I \in \mathcal{Z}. \tag{\ref{proj5-eq43}a}
	\end{align}
Since there are only limited feasible solutions of $I$, the optimal number of the aggregation rounds $I^*$ can be efficiently obtained by searching the solution space.
\subsection{The Whole Precedure of DP-OTA-FedAvg}\label{simulation}
In this subsection, we present the overall procedure of DP-OTA-FedAvg as shown in Algorithm 2 where $\mathcal{W}\left(\mathcal{K},\theta,I\right)= \eta ^{I}G+\frac{\varpi ^2}{\varrho} \left(1-\eta ^{I}\right) \left[4\left( 1-\frac{\left| \mathcal{K}\right|}{N} \right) ^2 +\left( \frac{T}{I}-1 \right) ^2 +\frac{1}{2}\frac{d\sigma ^2}{\left(\left| \mathcal{K} \right|\theta \right)^2}\right]$. It enables the BS to minimize the upper bound of the optimality gap within limited sum power and privacy budgets by designing the optimal device scheduling policy, alignment factor and the number of the aggregation rounds.

	\begin{algorithm}[ht]  
	\caption{Implementation of DP-OTA-FedAvg}  
	\label{Iteration}  
	\begin{algorithmic}[1]  
		\Require
		Given $N$, $d$, $\sigma$, $\left ( \epsilon,\xi  \right ) $, $\boldsymbol{h}=\left \{ \left | h_1\right |,\cdot \cdot \cdot,  \left | h_N\right |\right \}$, $P ^{dev}$ and $P ^{tot}$. 
		\State \underline{\emph{\textbf{Device scheduling}:}}
		\State Set the iteration number $j=0$ and $I^*=T$.
		\Repeat
		\State Obtain $\mathcal{K}^*_{\left(j\right)}$, $\theta ^*_{\left(j\right)}$ by Algorithm 1.
		\State Compute $I^*_{\left(j\right)}$ by solving P3.
		\State $j\gets j+1$.
		\Until the convergence condition $\left | \mathcal{W}\left (\mathcal{K}^*_{\left(j\right)}, \theta ^*_{\left(j\right)} ,I^*_{\left(j\right)}   \right ) -\mathcal{W}\left (\mathcal{K}^*_{\left(j-1\right)}, \theta ^*_{\left(j-1\right)} ,I^*_{\left(j-1\right)}   \right )  \right |\le\varepsilon  $ is satisfied;
		\State Obtain $\mathcal{K}^*= \mathcal{K}^*_{\left(j\right)}$, $\theta ^*=\theta ^*_{\left(j\right)}$, $I^*=I^*_{\left(j\right)}  $.
		
		\State \underline{\emph{\textbf{DP-OTA-FedAvg with limited power budget}:}}
		\State Set the initial communication round  $i=0$ and $\nu=\frac{\theta ^*}{\varpi }$.
		\For {$ i \in \left[ 0, I^*-1\right]$}
		\State The BS broadcasts the latest model $\boldsymbol{m}^i$ to the scheduled devices $\mathcal{K}^*$.
		\State Each device performs $E$ rounds of local training and computes the gradients by (\ref{proj5-eq3}), (\ref{proj5-eq4}), (\ref{proj5-eq5}).
		\State The gradients are transmitted and aggregated with the optimal alignment coefficient $\nu $ as shown in (\ref{proj5-eq9}).
		\State The BS recovers an average of the gradients by performing post-processing via (\ref{proj5-eq10}) and makes the update of the global model by (\ref{proj5-eq11}).
		\EndFor
		\Ensure Output $\boldsymbol{m}^{ I^*}$.
	\end{algorithmic}  
\end{algorithm}

\subsection{Optimial Device Scheduling and Alignment Factor with Different Peak Power} \label{proj5-subsectionE}
	With regard to the more general situation that each device has different transmit power budget $P_k$, we also define $\boldsymbol{c}=\left[ c_1,...c_m,...c_N \right]$ and $\boldsymbol{q}=\left[ q_1,...q_m,...q_N \right] $ where $c_m=\left | h_m \right | \sqrt{P_m} $ and $q_m=\sqrt{\frac{P^{tot}}{I}}\left ( 1/\sqrt{\sum_{j=m}^N\left ( 1/\left | h_j\right |^2 \right ) } \right ) $. The elements in $\boldsymbol{q}$ are sorted in ascending order because $\left | h_1\right |\le \left | h_2\right | \le ...\le \left | h_N\right |$ but the elements in $\boldsymbol{c}$ may not be. We define $\boldsymbol{c}^{sorted}=\left [ c_{1}^s,...,c_{m}^s,...,c_{N}^s \right ] $ as the list where the elements of $\boldsymbol{c}$ are sorted in the ascending order. 
Then, we can have similar results to the situation that each device has the same transmit power budget $P^{dev}$.

Similar to Lemma \ref{proj5-lemma4} we have the following result.
	\begin{lemma}\label{solution1}
		If $ \frac{\epsilon \sigma}{2\phi}<\min \left\{ c_1^s,q_1 \right\} $, the optimal solution to $\mathbf{P}2$ is
		\begin{align}\label{wg37}
			\theta =\frac{\epsilon \sigma}{2\phi},\quad \mathcal{K} =\mathcal{N}.
		\end{align}	 	
	\end{lemma} 
\itshape {Proof:}  \upshape The detailed proof is omitted as it is similar to the proof of Lemma \ref{proj5-lemma4}.
\hfill $\blacksquare$

Similarly, we define that $\mathcal{Q}^{diff} =\mathcal{Q}_1^{diff} \cup \mathcal{Q}_2^{diff}$ where $\mathcal{Q}_1^{diff} =\left\{ \left. k \right|c_1^s\le c_k^s< \frac{\epsilon \sigma}{2\phi} \right\}$ and $\mathcal{Q}_2^{diff} =\left\{ \left. k \right|q_1\le q_k< \frac{\epsilon \sigma}{2\phi} \right\}$. Then, we have the following Lemma similar to Lemma \ref{proj5-lemma5} .
	\begin{lemma}\label{proj5-lemma9}
		The minimum value of the potential optimal $\left| \mathcal{K} \right|$ is $N-\left| \mathcal{Q}^{diff}\right|$. The relationship between the potential optimal solution pairs, i.e.,  $\left| \mathcal{K} \right|$ and $\theta$ can be given by
		\begin{align}\label{wg37}
			\theta=
			\begin{cases} \min \left\{c_{N-\left| \mathcal{K} \right|+1}^s,q_{N-\left| \mathcal{K} \right|+1}\right\},if \left| \mathcal{K} \right| \ge N-\left| \mathcal{Q} ^{diff}\right|+1
				\\\frac{\epsilon \sigma}{2\phi}, if \left| \mathcal{K} \right| = N-\left| \mathcal{Q}^{diff} \right|
			\end{cases}.
		\end{align}
	\end{lemma} 
\itshape {Proof:}  \upshape The detailed proof is omitted as it is similar to the proof of Lemma \ref{proj5-lemma5}.
\hfill $\blacksquare$

Note that $\mathcal{Q}_2^{diff}$ is the same as $\mathcal{Q}_2$, which are independent of $P_k$. The difference between $\mathcal{Q}^{diff}$ and $\mathcal{Q}$ is caused by $\mathcal{Q}_1^{diff}$.

Following Lemma \ref{proj5-lemma9}, we can also derive the potential optimal solution pairs as follows.
	\begin{lemma}\label{closedsolution}
		There are $\left| \mathcal{Q} ^{diff}\right|+1$ closed-form solutions which may be the globally optimal solution. The $j$-th, $1\leq j \leq \left| \mathcal{Q}^{diff} \right|$, solution pair $\theta_j$ and $\mathcal{K} _j$ is given by
		\begin{align}\label{wg37}
			\theta_j = \min\left \{ c_j^s,q_j \right \} , \quad \mathcal{K} _j=\left\{\begin{matrix}\mathcal{K}_c, \text{if }c_j^s \le q_j
				
				\\
				\mathcal{K}_q ,\text{if }c_j^s > q_j
			\end{matrix}\right.
		\end{align}
		where $\mathcal{K}_c= \left\{ \left. k \right|c_k\ge c_j^s  \right\}$ and $\mathcal{K}_q =\left \{ k\mid k\ge j \right \} $.
	The $\left| \mathcal{Q}^{diff} \right|+1$-th solution pair $\theta_{\left| \mathcal{Q}^{diff} \right|+1}$, $\mathcal{K} _{\left| \mathcal{Q}^{diff} \right|+1}$ is 
		\begin{align}\label{wg37}
			\theta_ {\left| \mathcal{Q} ^{diff}\right|+1}= \frac{\epsilon \sigma}{2\phi}, \quad \mathcal{K} _{\left| \mathcal{Q}^{diff} \right|+1}=\left\{ \left. k \right|k \ge \left| \mathcal{Q}^{diff} \right|+1\right\}.
		\end{align}
	\end{lemma}
\itshape {Proof:}  \upshape The detailed proof is omitted as it is similar to the proof of Lemma \ref{proj5-lemma6}.
\hfill $\blacksquare$

Based on the above results, we can obtain the optimal solution to the problem of optimal device scheduling and alignment factor with different peak power by searching the space of the limited solution pairs.
	\section{Simulation Results}\label{proj5-section5}

	\subsection{Simulation Setting}\label{simulationsetting}
We evaluate our proposed scheme by training a convolutional neural network (CNN) on the popular MNIST \cite{lecun2010mnist} dataset used for handwritten digit classification. The MNIST dataset consists of 60,000 images for training and 10,000 testing images of the 10 digits.
We have the general assumption that there is an equal number of training data samples for each device and no overlap between the local training data sets \cite{zhu2019broadband} \cite{wang2019adaptive}.
We have the common assumption that each device has an equal amount of training data samples and the local training datasets are non-overlapping with each other \cite{zhu2019broadband} \cite{wang2019adaptive}. 
We assume that local datasets are IID, where the initial training dataset is randomly divided into $N$ batches and each device is assigned to one batch. In particular, CNN consists of two 5×5 convolution layers with the rectified linear unit (ReLU) activation. The two convolution layers have 10 and 20 channels respectively, and each layer has 2×2 max pooling, a fully-connected layer with 50 units and ReLU activation, and a log-softmax output layer, in which case $d= 21840$. The learning rate is set as $\eta=0.1$.
The peak transmission budget of each device is set to $1$ $W$. The number of total training rounds is $T=200$.
	
\subsection{Evaluation of Scheduling Policy }\label{performanceofBnb}
We first evaluate the performance of the proposed scheduling policy by comparing it with the uniform scheduling policy and the full device scheduling.
	\begin{figure*}[ht]
		\centering
		{  \begin{minipage}[t]{0.45\linewidth}
				\centering
				\includegraphics[scale=0.85]{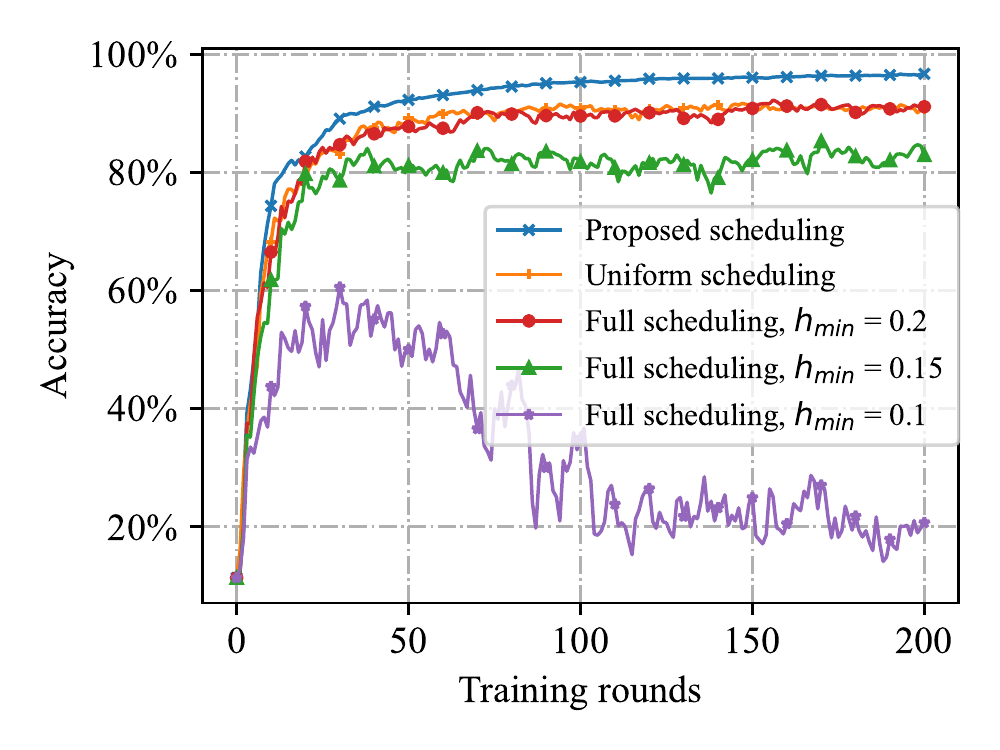}
			\end{minipage}	
		}
		{  \begin{minipage}[t]{0.45\linewidth}
				\centering
					\includegraphics[scale=0.85]{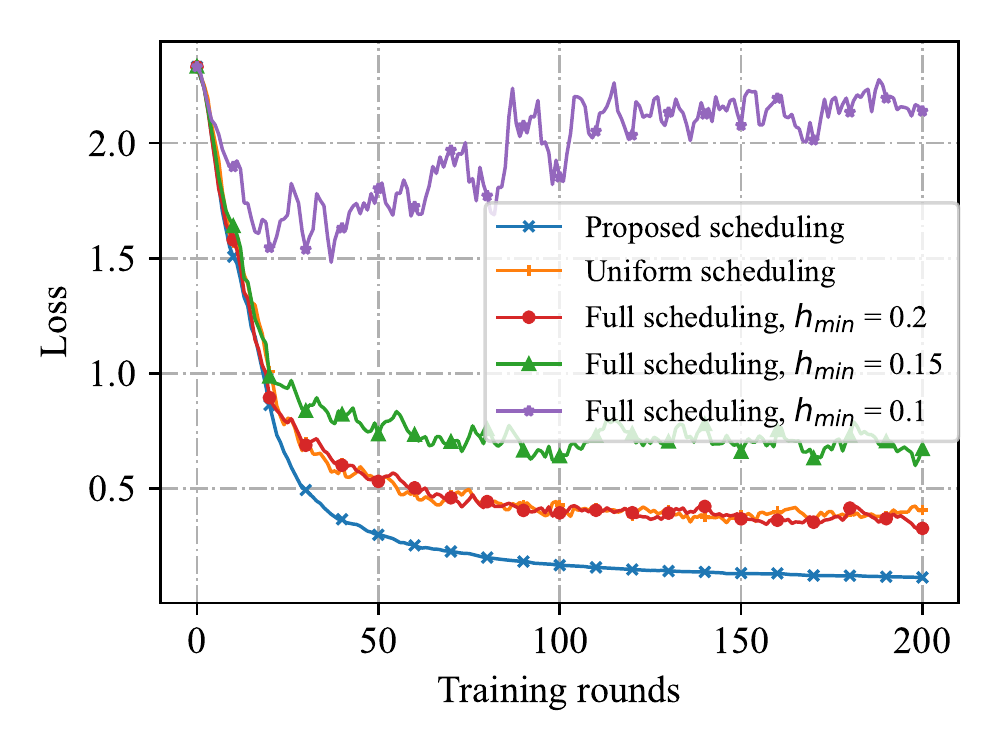}
			\end{minipage}	
		}
		\caption{The learning performance with different scheduling policies} \label{similation_scheduling}
	\end{figure*}
In Fig. \ref{similation_scheduling}, we plot the testing accuracy and training loss with different scheduling policies where the smallest channel coefficient of the proposed scheduling and uniform policies are all set to as $h_{min}=0.1$. This indicates that the proposed scheduling policy, the uniform scheduling, and the full scheduling with $h_{min}=0.1$ are performed under the same worst channel condition. It can be observed that the proposed scheduling performs better than uniform scheduling policies and full device scheduling. In particular, the superiority of the proposed scheduling policy is noticeable compared with the full scheduling scheme where some devices suffer from very poor channel conditions, i.e., $h_{min}=0.1$. This is because the alignment coefficient in the full scheduling policy is very small, which results in a quite low SNR of the DP-OTA-FedAvg system, degrading the utility of the aggregated gradient. Therefore, the proposed scheduling is especially useful for those FL networks where devices have poor channel conditions.

\subsection{Evaluation of The Impact of Aggregation Rounds}

\begin{figure*}[ht]
	\centering
	{  \begin{minipage}[t]{0.45\linewidth}
			\centering
			\includegraphics[scale=0.85]{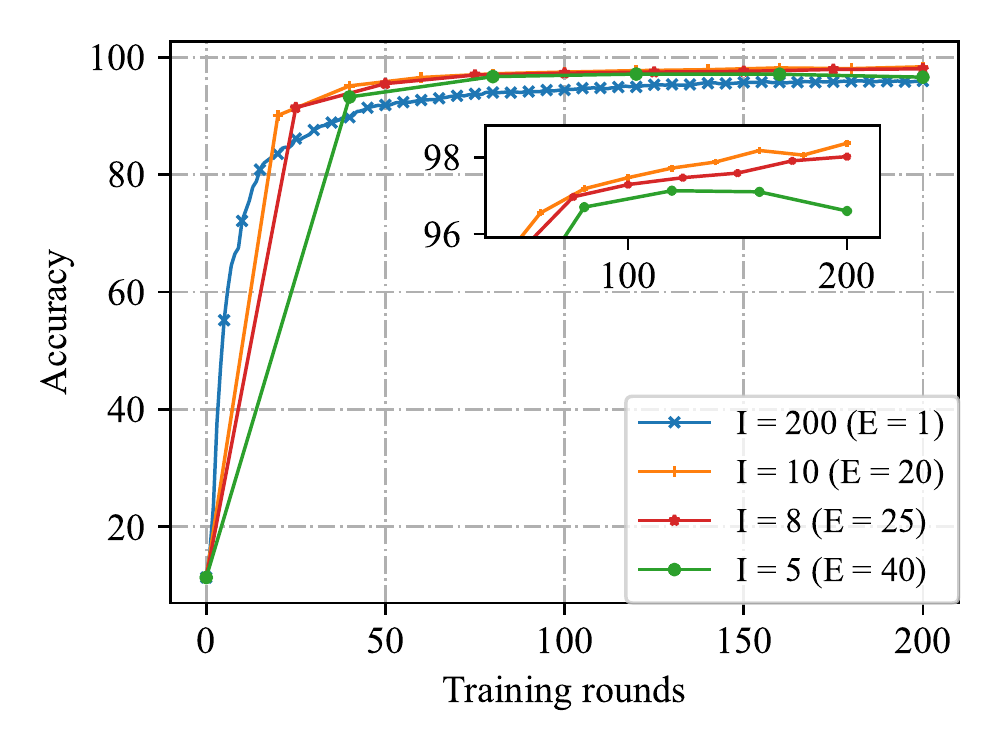}
		\end{minipage}	
	}
	{  \begin{minipage}[t]{0.45\linewidth}
			\centering
			\includegraphics[scale=0.85]{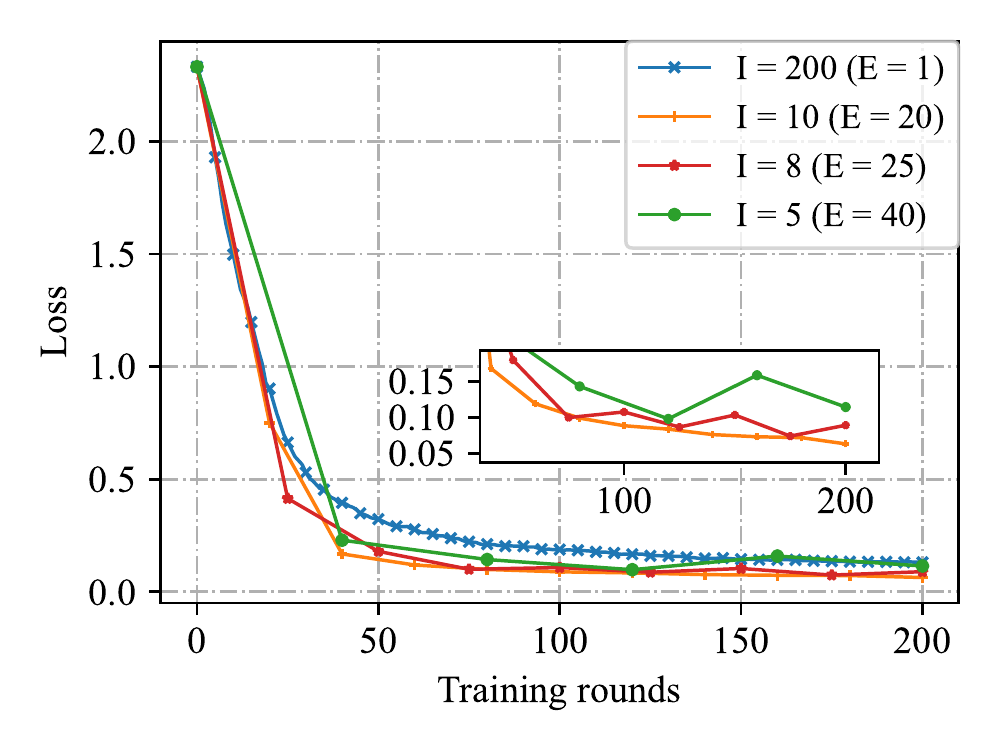}
		\end{minipage}	
	}
	\caption{The learning performance with different number of aggregation rounds} \label{simulation_localtimes}
\end{figure*}

\begin{figure*}[ht]
	\centering
	{  \begin{minipage}[t]{0.45\linewidth}
			\centering
			\includegraphics[scale=0.9]{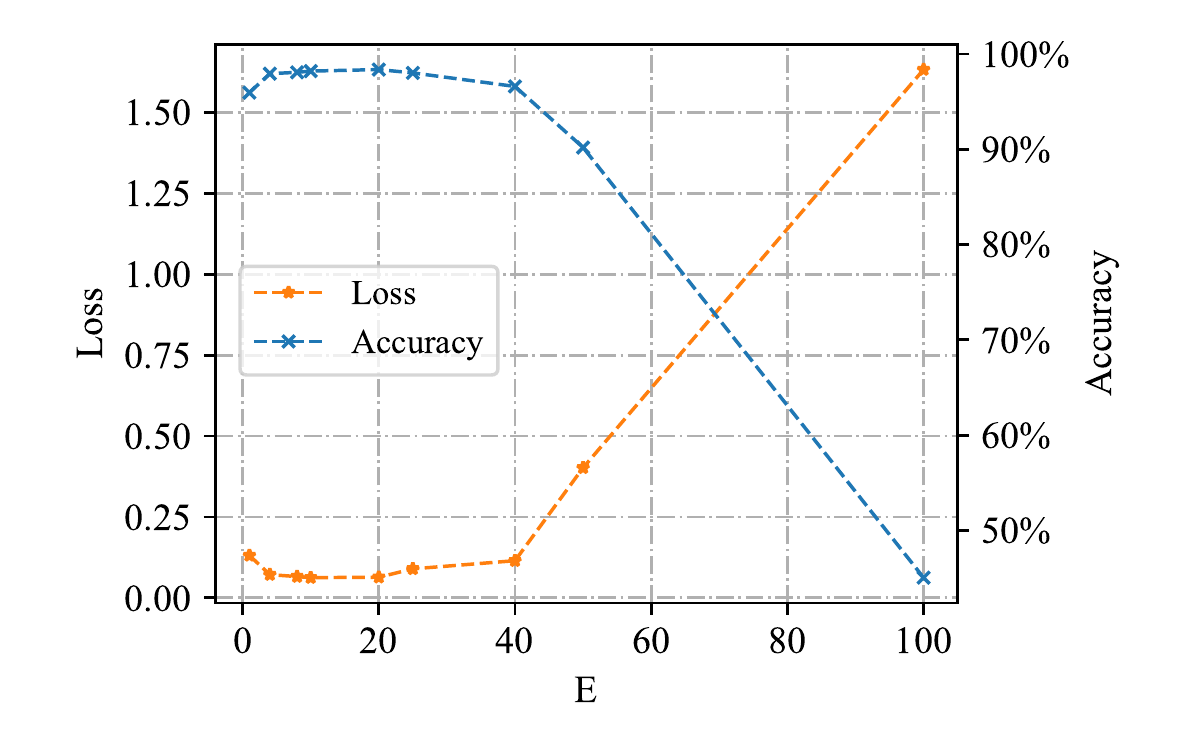}
		\end{minipage}	
	}
	\caption{The accuracy and loss with local training times $E$} \label{simulation_localtimes_trad}
\end{figure*}

In Fig. \ref{simulation_localtimes}, we plot the testing accuracy and training loss with the different numbers of aggregation rounds given the total training rounds $T = 200$. The learning performance is first improved and then degraded as $I$ decreases, i.e, $E$ increases. It can be understood intuitively that a smaller $I$ can reduce the distortion caused by the transmission during the aggregation. However, if $I$ is set too small, which means a large $E$, the local training may converge to the optimal solution of the local objective rather than the global objective. Therefore, there exists an optimal number of local training rounds $E$ to balance the communication distortion and the local training error as shown in Fig. \ref{simulation_localtimes_trad}.

\subsection{Evaluation of The Optimal Design of DP-OTA-FedAvg}
We evaluate the performance of the proposed overall scheme of DP-OTA-FedAvg where $h_{min}=0.2$, $P^{tot}= 1000W$, $N = 100$.
	\begin{figure*}[ht]
		\centering
		{  \begin{minipage}[t]{0.45\linewidth}
				\centering
				\includegraphics[scale=0.85]{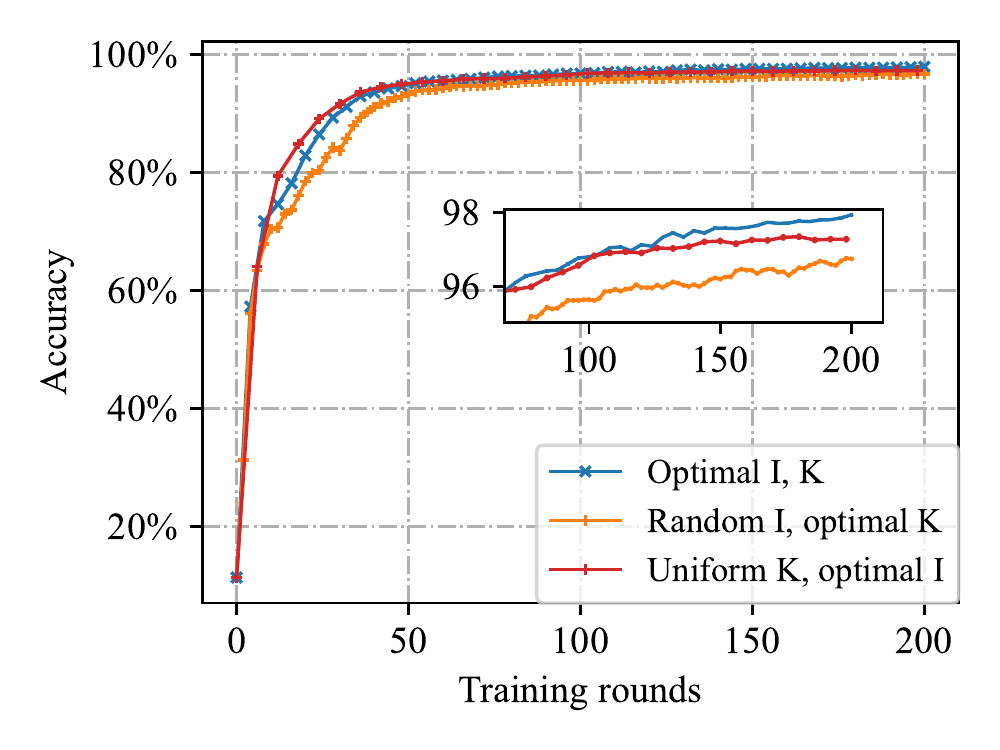}
			\end{minipage}	
		}
		{  \begin{minipage}[t]{0.45\linewidth}
				\centering
				\includegraphics[scale=0.85]{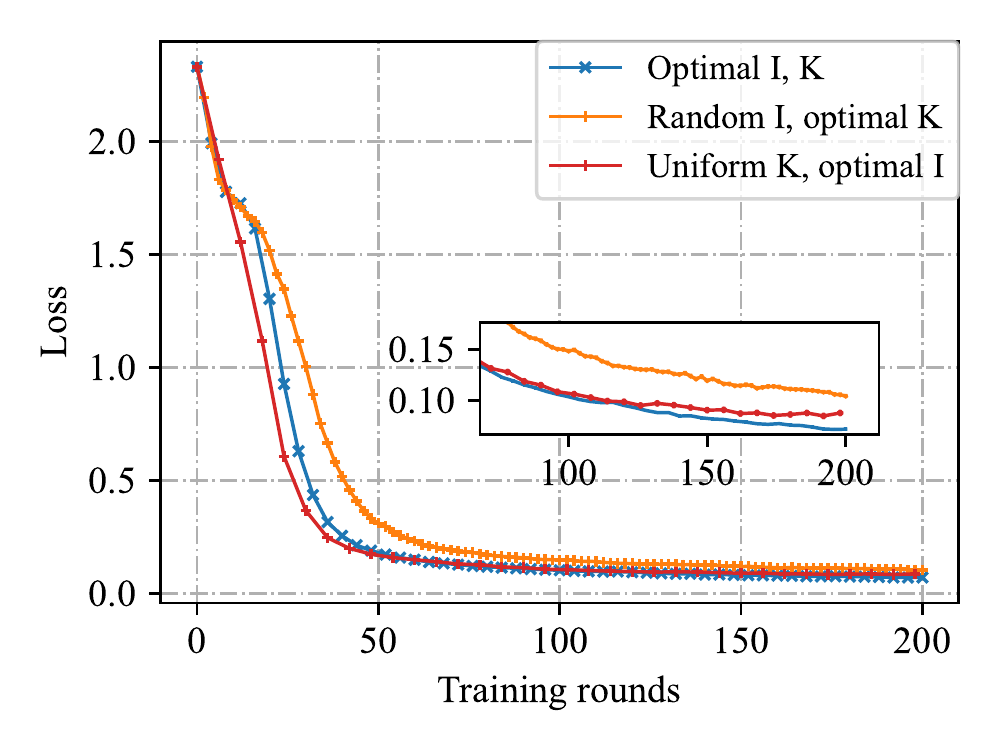}
			\end{minipage}	
		}		
		\caption{The learning performance with optimal design} \label{simulation_overall}
	\end{figure*}
Fig. \ref{simulation_overall} shows the learning accuracy and loss of the optimal design of device scheduling and the number of the aggregation rounds. It demonstrates that the proposed optimal design can significantly improve the performance of DP-OTA-FedAvg. On the one hand, the optimal device scheduling and alignment coefficient design enhances the learning performance by improving the utility of the aggregated gradient average in each communication round. On the other hand, DP-OTA-FedAvg can set more efficient local training rounds by optimizing the number of the global model aggregations under the constraint of limited sum power and privacy budget.
\section{Conclusion}\label{proj5-section6}
This paper has studied the optimal design of device scheduling, alignment coefficient, and the number of communication rounds of DP-OTA-FedAvg with limited sum power and privacy budgets. 
The privacy analysis has shown that a smaller alignment coefficient is beneficial for privacy preservation while having a negative impact on the learning performance according to the convergence analysis. However, there is a tradeoff between the number of the scheduled device and the alignment coefficient. Additionally, the design of device scheduling and alignment coefficient also affects the number of aggregation rounds of DP-OTA-FedAvg with limited sum power. To this end, we have formulated an optimization problem to minimize the optimality gap considering the privacy and sum power constraints. We have obtained the closed-form expression of the relationship between the number of scheduled devices and the alignment coefficient, which offered limited pairs of the potential optimal solution. Then, the optimal solutions were obtained efficiently by searching a limited solution space. 

	\vspace{-15pt}


	
	
	\appendices
	\section{Proof of Lemma  \ref{proj5-lemma2}}\label{proj5-lemma2-proof}
	Following (\ref{proj5-eq5}), (\ref{proj5-eq12}) and \ref{proj5-eq13}, we have
	\begin{equation}\label{alignmentcoefficient}
		\begin{aligned}
			\boldsymbol{m}^{i+1}-\boldsymbol{m}^{i} & = -\tau \boldsymbol{\tilde{g}}^i = -\tau\left ( \frac{1}{\left| \mathcal{K} \right|}\sum_{k\in \mathcal{K}}{\boldsymbol{g}_{k}^{i}}+\frac{1}{\left| \mathcal{K} \right|\nu }\boldsymbol{r}^i\right )
			\\
			&=-\tau\left ( \frac{1}{\left| \mathcal{K} \right|}\sum_{k\in \mathcal{K}}{ \sum_{\iota = 0}^{E-1} \nabla L_k\left( \boldsymbol{w}_{k}^{i,\iota} \right) }+\frac{1}{\left| \mathcal{K} \right|\nu }\boldsymbol{r}^i\right ).
		\end{aligned}
	\end{equation}
	Then, we have
		\begin{align} \label{proj5-eq49}
			&\mathbb{E} \left [ L\left ( \boldsymbol{m}^{i+1}  \right )  \right ] -\mathbb{E} \left [ L\left ( \boldsymbol{m}^{i}  \right )  \right ]
			\\
			\overset{\left( a \right)}{\le}& \mathbb{E}\left [ \left \langle \nabla L \left ( \boldsymbol{m}^{i}  \right ) ,\boldsymbol{m}^{i+1}-\boldsymbol{m}^{i} \right \rangle  \right ]+\frac{\zeta }{2}\mathbb{E} \left [ \left \| \boldsymbol{m}^{i+1}-\boldsymbol{m}^{i} \right \|_{2}^{2}  \right ]    \nonumber
			\\
			=&-\tau \mathbb{E}\left [ \left \langle \nabla L \left ( \boldsymbol{m}^{i}  \right ) ,\frac{1}{\left| \mathcal{K} \right|}\sum_{k\in \mathcal{K}}{\boldsymbol{g}_{k}^{i}}+\frac{1}{\left| \mathcal{K} \right|\nu }\boldsymbol{r}^i \right \rangle  \right ]+\frac{\zeta \tau^2}{2}\mathbb{E} \left [ \left \|\frac{1}{\left| \mathcal{K} \right|}\sum_{k\in \mathcal{K}}{\boldsymbol{g}_{k}^{i}}+\frac{1}{\left| \mathcal{K} \right|\nu }\boldsymbol{r}^i \right \|_{2}^{2}  \right ]    
			\nonumber \\
			=&-\tau \mathbb{E}\left [ \left \langle \nabla L \left ( \boldsymbol{m}^{i}  \right ) ,{\frac{1}{\left| \mathcal{K} \right|}\sum_{k\in \mathcal{K}}{ \sum_{\iota = 0}^{E-1} \nabla L_k\left( \boldsymbol{w}_{k}^{i,\iota} \right) }}\right \rangle  \right ]
			- \frac{\tau}{\left| \mathcal{K} \right|\nu } \left \langle \nabla L \left ( \boldsymbol{m}^{i}  \right ) ,\mathbb{E}\left [\boldsymbol{r}^i   \right ] \right \rangle
			\nonumber \\
			&+\frac{\zeta \tau^2}{2}\mathbb{E} \left [ \left \|{\frac{1}{\left| \mathcal{K} \right|}\sum_{k\in \mathcal{K}}{ \sum_{\iota = 0}^{E-1} \nabla L_k\left( \boldsymbol{w}_{k}^{i,\iota} \right) }} \right \|_{2}^{2}  \right ]
			+\frac{\zeta \tau^2}{2}\mathbb{E} \left [ \left \|\frac{1}{\left| \mathcal{K} \right|\nu }\boldsymbol{r}^i \right \|_{2}^{2}  \right ]    
			\nonumber \\
			&+\frac{ \zeta \tau^2}{\left| \mathcal{K} \right|\nu }  \left \langle {\frac{1}{\left| \mathcal{K} \right|}\sum_{k\in \mathcal{K}}{ \sum_{\iota = 0}^{E-1} \nabla L_k\left( \boldsymbol{w}_{k}^{i,\iota} \right) }} ,\mathbb{E}\left [ \boldsymbol{r}^i   \right ]\right \rangle
	     	\nonumber	\\
			\overset{\left( b \right)}{=}&\underset{A}{\underbrace{-\tau \mathbb{E} \left[ \left. \langle \nabla L\left( \boldsymbol{m}^i \right) ,\frac{1}{\left| \mathcal{K} \right|}\sum_{k\in \mathcal{K}}{\sum_{\iota =0}^{E-1}{\nabla L_k\left( \boldsymbol{w}_{k}^{i,\iota} \right)}} \right. \rangle \right] }}
	    	\nonumber	\\
			&+\underset{B}{\underbrace{\frac{\zeta \tau ^2}{2}\mathbb{E} \left[ \left\| \frac{1}{\left| \mathcal{K} \right|}\sum_{k\in \mathcal{K}}{\sum_{\iota =0}^{E-1}{\nabla}L_k\left( \boldsymbol{w}_{k}^{i,\iota} \right)} \right\| _{2}^{2} \right] }}+\underset{C}{\underbrace{\frac{\zeta \tau ^2}{2}\mathbb{E} \left[ \left\| \frac{1}{\left| \mathcal{K} \right|\nu}\boldsymbol{r}^i \right\| _{2}^{2} \right] }}	\nonumber,
		\end{align}
	where (a) is from Assumption \ref{proj5-ass2} and (b) is come from the fact that $\mathbb{E}\left [ \boldsymbol{r}^i   \right ]=0$.
	To obtain the upper bound of term $A$,  term $A$ is rewritten as follows
		\begin{align}
			A&=-\tau \mathbb{E} \left[ \left. \langle \nabla L\left( \boldsymbol{m}^i \right) ,\nabla L\left( \boldsymbol{m}^i \right) +\frac{1}{\left| \mathcal{K} \right|}\sum_{k\in \mathcal{K}}{\sum_{\iota =0}^{E-1}{\nabla L_k\left( \boldsymbol{w}_{k}^{i,\iota} \right)}-\nabla L\left( \boldsymbol{m}^i \right)} \right. \rangle \right] 
		\\
			=&-\tau \mathbb{E} \left[ \left\| \nabla L\left( \boldsymbol{m}^i \right) \right\| _{2}^{2} \right] +\tau \mathbb{E} \left[ \left. \langle \nabla L\left( \boldsymbol{m}^i \right) ,\nabla L\left( \boldsymbol{m}^i \right) -\frac{1}{\left| \mathcal{K} \right|}\sum_{k\in \mathcal{K}}{\sum_{\iota =0}^{E-1}{\nabla L_k\left( \boldsymbol{w}_{k}^{i,\iota} \right)}} \right. \rangle \right] 
		\nonumber	\\
			=&-\tau \mathbb{E} \left[ \left\| \nabla L\left( \boldsymbol{m}^i \right) \right\| _{2}^{2} \right] +\frac{\tau}{2}\mathbb{E} \left[ \left\| \nabla L\left( \boldsymbol{m}^i \right) \right\| _{2}^{2} \right] 
		\nonumber	\\
			&+\frac{\tau}{2}\mathbb{E} \left[ \left\| \nabla L\left( \boldsymbol{m}^i \right) -\frac{1}{\left| \mathcal{K} \right|}\sum_{k\in \mathcal{K}}{\sum_{\iota =0}^{E-1}{\nabla L_k\left( \boldsymbol{w}_{k}^{i,\iota} \right)}} \right\| _{2}^{2} \right] -\frac{\tau}{2}\mathbb{E} \left[ \left\| \frac{1}{\left| \mathcal{K} \right|}\sum_{k\in \mathcal{K}}{\sum_{\iota =0}^{E-1}{\nabla L_k\left( \boldsymbol{w}_{k}^{i,\iota} \right)}} \right\| _{2}^{2} \right] 
		\nonumber	\\
			=&-\frac{\tau}{2}\mathbb{E} \left[ \left\| \nabla L\left( \boldsymbol{m}^i \right) \right\| _{2}^{2} \right] +\underset{A_1}{\underbrace{\frac{\tau}{2}\mathbb{E} \left[ \left\| \nabla L\left( \boldsymbol{m}^i \right) -\frac{1}{\left| \mathcal{K} \right|}\sum_{k\in \mathcal{K}}{\sum_{\iota =0}^{E-1}{\nabla L_k\left( \boldsymbol{w}_{k}^{i,\iota} \right)}} \right\| _{2}^{2} \right] }}
			\nonumber\\
			&\underset{A_2}{\underbrace{-\frac{\tau}{2}\mathbb{E} \left[ \left\| \frac{1}{\left| \mathcal{K} \right|}\sum_{k\in \mathcal{K}}{\sum_{\iota =0}^{E-1}{\nabla L_k\left( \boldsymbol{w}_{k}^{i,\iota} \right)}} \right\| _{2}^{2} \right] }}\nonumber.
		\end{align}
	The upper bound of term $A_1$ is obtained as follows
	\begin{align}
			A_1&=\frac{\tau}{2}\mathbb{E} \left[ \left\| \frac{1}{N}\sum_{k\in \mathcal{N}}{\nabla L_k\left( \boldsymbol{m}^i \right)}-\frac{1}{\left| \mathcal{K} \right|}\sum_{k\in \mathcal{K}}{\nabla L_k\left( \boldsymbol{m}^i \right)} \right. \right. 
	\\
	&	\left. \left. +\frac{1}{\left| \mathcal{K} \right|}\sum_{k\in \mathcal{K}}{\nabla L_k\left( \boldsymbol{m}^i \right)}-\frac{1}{\left| \mathcal{K} \right|}\sum_{k\in \mathcal{K}}{\sum_{\iota =0}^{E-1}{\nabla L_k\left( \boldsymbol{w}_{k}^{i,\iota} \right)}} \right\| _{2}^{2} \right] 
		\nonumber 	\\
			=&\frac{\tau}{2}\mathbb{E} \left[ \left\| \left( \frac{1}{N}-\frac{1}{\left| \mathcal{K} \right|} \right) \sum_{k\in \mathcal{K}}{\nabla L_k\left( \boldsymbol{m}^i \right)}+\frac{1}{N}\sum_{k\in \mathcal{N} /\mathcal{K}}{\nabla L_k\left( \boldsymbol{m}^i \right)} \right. \right. 
		\nonumber 	\\
		&	\left. \left. 
			+\frac{1}{\left| \mathcal{K} \right|}\sum_{k\in \mathcal{K}}{\left( \nabla L_k\left( \boldsymbol{m}^i \right) -\sum_{\iota =0}^{E-1}{\nabla L_k\left( \boldsymbol{w}_{k}^{i,\iota} \right)} \right)} \right\| _{2}^{2} \right] 
		\nonumber 	\\
			\overset{\left( a \right)}{\le}&\tau\mathbb{E} \left[ \left\| \left( \frac{1}{N}-\frac{1}{\left| \mathcal{K} \right|} \right) \sum_{k\in \mathcal{K}}{\nabla L_k\left( \boldsymbol{m}^i \right)}+\frac{1}{N}\sum_{k\in \mathcal{N} /\mathcal{K}}{\nabla L_k\left( \boldsymbol{m}^i \right)} \right\| _{2}^{2} \right] 
		\nonumber 	\\
			&+\tau\mathbb{E} \left[ \left\| \frac{1}{\left| \mathcal{K} \right|}\sum_{k\in \mathcal{K}}{\left( \nabla L_k\left( \boldsymbol{m}^i \right) -\sum_{\iota =0}^{E-1}{\nabla L_k\left( \boldsymbol{w}_{k}^{i,\iota} \right)} \right)} \right\| _{2}^{2} \right] 
		\nonumber 	\\
			\overset{\left( b \right)}{\le}&\tau\left[\left( \frac{1}{\left| \mathcal{K} \right|}-\frac{1}{N} \right) \sum_{k\in \mathcal{K}}{\mathbb{E} \left[ \left\| \nabla L_k\left( \boldsymbol{m}^i \right) \right\| _2 \right]}+\frac{1}{N}\sum_{k\in \mathcal{N} /\mathcal{K}}{\mathbb{E} \left[ \left\| \nabla L_k\left( \boldsymbol{m}^i \right) \right\| _2 \right]} \right] ^2
		\nonumber 	\\
			&+\tau\left( \frac{1}{\left| \mathcal{K} \right|}\sum_{k\in \mathcal{K}}{\sum_{\iota =1}^{E-1}{\mathbb{E} \left[ \left\| \nabla L_k\left( \boldsymbol{w}_{k}^{i,\iota} \right) \right\| _2 \right]}} \right) ^2
		\nonumber 	\\
			\overset{\left( c \right)}{\le}&4\tau\varpi ^2\left( 1-\frac{\left| \mathcal{K} \right|}{N} \right) ^2 +\tau \varpi ^2\left( E-1 \right) ^2 \nonumber ,
	\end{align}
	where (a) is from that $\left \| a+b \right \| _{2}^{2} \le 2\left \| a\right \| _{2}^{2}+2\left \| b \right \| _{2}^{2}$ and (b) is from $\left \| a+b+c \right \| _{2}^{2} \le \left (  \left \| a\right \| _{2}+\left \| b\right \| _{2}+\left \| c\right \| _{2}\right )^2 $. Inequality (c) comes from Assumption \ref{proj5-ass1}.
	Due to $ \tau \le\frac{1}{\zeta} $, we obtain the upper bound of the sum of term $A_2$ and term $B$ as follows
	\begin{equation}
		\begin{aligned}
			A_2+B=\frac{\tau}{2}\left( \zeta \tau -1 \right) \mathbb{E} \left[ \left\| \frac{1}{\left| \mathcal{K} \right|}\sum_{k\in \mathcal{K}}{\sum_{\iota =0}^{E-1}{\nabla L_k\left( \boldsymbol{w}_{k}^{i,\iota} \right)}} \right\| _{2}^{2} \right]\le0.
		\end{aligned}
	\end{equation}
	For the last term $C$, we note that
	\begin{equation}
		\begin{aligned}
			C=\frac{\zeta \tau ^2}{2}\mathbb{E} \left[ \left\| \frac{1}{\left| \mathcal{K} \right|\nu}\boldsymbol{r}^i \right\| _{2}^{2} \right] =\frac{\zeta \tau ^2}{2}\frac{d\sigma ^2}{\left| \mathcal{K} \right|^2\nu ^2}.
		\end{aligned}
	\end{equation}
	By pluggling these upper bounds back into (\ref{proj5-eq49}), we complete the proof as follows:
	\begin{equation}
		\begin{aligned}
			&\mathbb{E} \left[ L\left( \boldsymbol{m}^{i+1} \right) \right]-\mathbb{E} \left[ L\left( \boldsymbol{m}^{i} \right) \right]  
			\\
			\le& -\frac{\tau}{2}\mathbb{E} \left[ \left\| \nabla L\left( \boldsymbol{m}^i \right) \right\| _{2}^{2} \right] +4\tau\varpi ^2\left( 1-\frac{\left| \mathcal{K} \right|}{N} \right) ^2 +\tau \varpi ^2\left( E-1 \right) ^2 +\frac{\zeta \tau ^2}{2}\frac{d\sigma ^2}{\left| \mathcal{K} \right|^2\nu ^2}.
			\\
		\end{aligned}
	\end{equation}
\section{Proof of Theorem \ref{proj5-theorem1}}\label{proj5-theorem1-proof}
	Based on Lemma \ref{proj5-lemma2} and Assumption \ref{proj5-ass3}, we have 	
		\begin{align}
			&\mathbb{E} \left[ L\left( \boldsymbol{m}^{i+1} \right) \right] -\mathbb{E} \left[ L\left( \boldsymbol{m}^* \right) \right] \\
			\overset{\left( a \right)}{\le}&\eta\left[ \mathbb{E} \left[ L\left( \boldsymbol{m}^i \right) \right] -\mathbb{E} \left[ L\left( \boldsymbol{m}^* \right) \right] \right] +\frac{\varpi ^2}{\zeta}\left[ 4\left( 1-\frac{\left| \mathcal{K} \right|}{N} \right) ^2+\left( E-1 \right) ^2+\frac{d\sigma ^2}{2\left| \mathcal{K} \right|^2\theta ^2} \right] 
			\nonumber\\
			=&\eta^{i+1}\left[ \mathbb{E} \left[ L\left( \boldsymbol{m}^0 \right) \right] -\mathbb{E} \left[ L\left( \boldsymbol{m}^* \right) \right] \right] +\frac{\varpi ^2}{\zeta}\left[ 4\left( 1-\frac{\left| \mathcal{K} \right|}{N} \right) ^2+\left( E-1 \right) ^2+\frac{d\sigma ^2}{2\left| \mathcal{K} \right|^2\theta ^2} \right] \sum_{\kappa =0}^{i}{\eta^{\kappa}}
			\nonumber
			\\
			=&\eta^{i+1}\left[ \mathbb{E} \left[ L\left( \boldsymbol{m}^0 \right) \right] -\mathbb{E} \left[ L\left( \boldsymbol{m}^* \right) \right] \right] +\frac{\varpi ^2}{\varrho}\left(1-\eta ^{i+1}\right)\left[ 4\left( 1-\frac{\left| \mathcal{K} \right|}{N} \right) ^2+\left( E-1 \right) ^2+\frac{d\sigma ^2}{2\left| \mathcal{K} \right|^2\theta ^2} \right] ,\nonumber
		\end{align}
	where $\eta =1-\frac{\varrho}{\zeta}$ and (a) is from (\ref{proj5-eq27}). By replacing $i+1$ with $I$, we complete the proof.
	
	\section{Proof of Theorem \ref{proj5-theorem2}}\label{proj5-theorem2-proof}
	Based on Lemma \ref{proj5-lemma2}, we have 
		\begin{equation}
		\begin{aligned}
			&\mathbb{E} \left[ \left\| \nabla L\left( \boldsymbol{m}^i \right) \right\| _{2}^{2} \right] 
			\\
			\le &\frac{2}{\tau}\left [\mathbb{E} \left[ L\left( \boldsymbol{m}^{i+1} \right) \right] - \mathbb{E} \left[ L\left( \boldsymbol{m}^{i} \right) \right] 	+4\tau\varpi ^2\left( 1-\frac{\left| \mathcal{K} \right|}{N} \right) ^2 
			+\tau \varpi ^2\left( E-1 \right) ^2 +\frac{\zeta \tau ^2}{2}\frac{d\sigma ^2}{\left| \mathcal{K} \right|^2\nu ^2} \right ]
			\\
			=&\frac{2}{\tau}\left [\mathbb{E} \left[ L\left( \boldsymbol{m}^{i+1} \right) \right] - \mathbb{E} \left[ L\left( \boldsymbol{m}^{i} \right) \right] \right ]+\varpi ^2\left[ 8\left( 1-\frac{\left| \mathcal{K} \right|}{N} \right) ^2+2\left( E-1 \right) ^2+\frac{d\sigma ^2}{\left| \mathcal{K} \right|^2\theta ^2} \right],
		\end{aligned}
	\end{equation}
	where (a) is from $\tau=\frac{1}{\zeta}$.
	By summing $i$ from 0 to $I-1$, we complete the proof of Theorem \ref{proj5-theorem2} as follows:
		\begin{align}
			&\frac{1}{I}\sum_{i=0}^{I-1}{\mathbb{E} \left[ \left\| \nabla L\left( \boldsymbol{m}^i \right) \right\| _{2}^{2} \right]}
					\\
					\le& \frac{2}{\tau I}\left[ \mathbb{E} \left[ L\left( \boldsymbol{m}^0 \right) \right] -\left[ L\left( \boldsymbol{m}^I \right) \right] \right] +\varpi ^2\left[ 8\left( 1-\frac{\left| \mathcal{K} \right|}{N} \right) ^2+2\left( E-1 \right) ^2+\frac{d\sigma ^2}{\left| \mathcal{K} \right|^2\theta ^2} \right] \nonumber 
				\\
						\overset{\left( a \right)}{\le}& \frac{2}{\tau I}\left[ \mathbb{E} \left[ L\left( \boldsymbol{m}^0 \right) \right] -\left[ L\left( \boldsymbol{m}^* \right) \right] \right] +\varpi ^2\left[ 8\left( 1-\frac{\left| \mathcal{K} \right|}{N} \right) ^2+2\left( E-1 \right) ^2+\frac{d\sigma ^2}{\left| \mathcal{K} \right|^2\theta ^2} \right],\nonumber 
		\end{align}
	where (a) comes from the fact that $L\left( \boldsymbol{m}^* \right)\le L\left( \boldsymbol{m}^I \right)$.

	\bibliographystyle{IEEEtran}
	\bibliography{FLbib}

	\end{document}